\documentclass[10pt,twocolumn,letterpaper]{article}

\usepackage{cvpr}              %
\usepackage[accsupp]{axessibility}

\usepackage{xspace}
\usepackage{url}
\usepackage{graphics} %

\usepackage{graphicx}
\usepackage{gensymb}
\usepackage{amsmath}
\usepackage{multirow}
\usepackage{makecell}
\usepackage{booktabs}
\usepackage{afterpage}
\usepackage{caption}
\usepackage{wrapfig}

\def\reprFull{Scene Language\@\xspace}

\DeclareMathOperator*{\argmin}{arg\,min}

\makeatletter
\DeclareRobustCommand\onedot{\futurelet\@let@token\@onedot}
\def\@onedot{\ifx\@let@token.\else.\null\fi\xspace}

\def\eg{\emph{e.g}\onedot} 
\def\ie{\emph{i.e}\onedot}

\makeatother

\newcommand{\xpm}[1]{{\tiny$\pm#1$}}

\definecolor{myProgColor}{RGB}{233,113,50}  %
\definecolor{myWordColor}{RGB}{21,96,130}  %
\definecolor{myEmbdColor}{RGB}{160,43,147}   %
\def\prog{\textcolor{myProgColor}{program}\xspace}
\def\progs{\textcolor{myProgColor}{programs}\xspace}

\def\progVar{\textcolor{myProgColor}{program}\xspace}
\def\word{\textcolor{myWordColor}{word}\xspace}
\def\words{\textcolor{myWordColor}{words}\xspace}
\def\Word{\textcolor{myWordColor}{Word}\xspace}

\def\WordType{\textcolor{myWordColor}{Word}\xspace}

\def\embd{\textcolor{myEmbdColor}{embedding}\xspace}
\def\embds{\textcolor{myEmbdColor}{embeddings}\xspace}
\def\Embd{\textcolor{myEmbdColor}{Embedding}\xspace}

\def\EmbdType{\textcolor{myEmbdColor}{Embedding}\xspace}

\definecolor{codegreen}{rgb}{0,0.6,0}
\definecolor{codegray}{rgb}{0.5,0.5,0.5}
\definecolor{codepurple}{rgb}{0.58,0,0.82}
\definecolor{backcolour}{rgb}{0.95,0.95,0.92}

\usepackage[frozencache=true,cachedir=minted-cache]{minted}
\setminted{
  fontsize=\scriptsize,
  baselinestretch=.9,
  breaklines=true,
  breakanywhere=true,
  funcnamehighlighting=true,
  tabsize=4,
  obeytabs=false,
  mathescape=false,
  samepage=false,
  showspaces=false,
  showtabs=false,
  texcl=false,
}

\usepackage{mdframed}
\newcommand{\styledmarkdowninput}[1]{%
  \begin{mdframed}[backgroundcolor=gray!10,linewidth=0pt]
      {\ttfamily\scriptsize\inputminted{markdown}{#1}}
  \end{mdframed}
}

\renewcommand{\paragraph}[1]{\vspace{0.1cm}\noindent\textbf{#1}}

\definecolor{cvprblue}{rgb}{0.21,0.49,0.74}
\usepackage[pagebackref,breaklinks,colorlinks,allcolors=cvprblue]{hyperref}

\usepackage[capitalize]{cleveref}
\usepackage[size=small]{caption}

\def\paperID{2228} %
\def\confName{CVPR}
\def\confYear{2025}

\title{The Scene Language: Representing Scenes\\ with Programs, Words, and Embeddings}

\author{
Yunzhi Zhang$^{1}$ \quad
Zizhang Li$^{1}$ \quad
Matt Zhou$^{2}$ \quad
Shangzhe Wu$^{1}$ \quad
Jiajun Wu$^{1}$ \\[0.4em]
$^1$Stanford University \quad $^2$UC Berkeley
}

\begin{document}

\twocolumn[
    \maketitle
\begin{center}
\vspace{-1.8em}
    \includegraphics[trim={0pt 0pt 0pt 5pt}, clip, width=\textwidth]{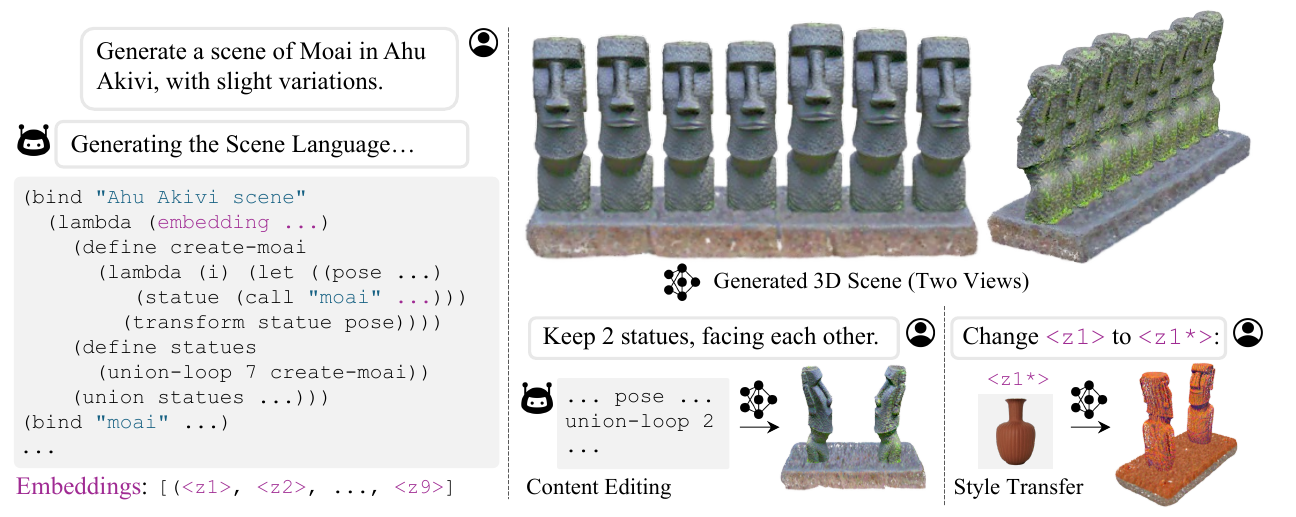}
    \vspace{-2.3em}
    \captionof{figure}{\textbf{Structured Scene Generation and Editing Using the \reprFull.}
    We develop a scene representation for 3D scene generation and editing tasks.
    Given textual scene descriptions, the representation can be inferred by a pre-trained large language model, rendered in 3D, and edited following language instructions. 
    The representation contains a \prog consisting of semantic-aware functions bound to \words, providing high interpretability and an intuitive scene-editing interface, and \embds enabling editing with fine controls, \eg, transferring the style of \texttt{\textcolor{myEmbdColor}{<z1*>}} from a user's image to the output by updating \texttt{\textcolor{myEmbdColor}{<z1>}} which controls global attributes of the scene.
    Project page: {\footnotesize\url{https://ai.stanford.edu/~yzzhang/projects/scene-language/}}.
    }    
    \vspace{-0.4em}
    \label{fig:teaser}
\end{center}
    \bigbreak
]

\begin{abstract}
We introduce the \reprFull, a visual scene representation that concisely and precisely describes the structure, semantics, and identity of visual scenes.
It represents a scene with three key components: a \prog that specifies the hierarchical and relational structure of entities in the scene,
\words in natural language that summarize the semantic class of each entity, and
\embds that capture the visual identity of each entity.
This representation can be inferred from pre-trained language models via a training-free inference technique, given text or image inputs.
The resulting scene can be rendered into images using traditional, neural, or hybrid graphics renderers.
Together, this forms an automated system for high-quality 3D and 4D scene generation.
Compared with existing representations like scene graphs, our proposed \reprFull generates complex scenes with higher fidelity, while explicitly modeling the scene structures to enable precise control and editing.

\end{abstract}

\begin{figure*}[t]
    \centering
    \includegraphics[width=\textwidth]{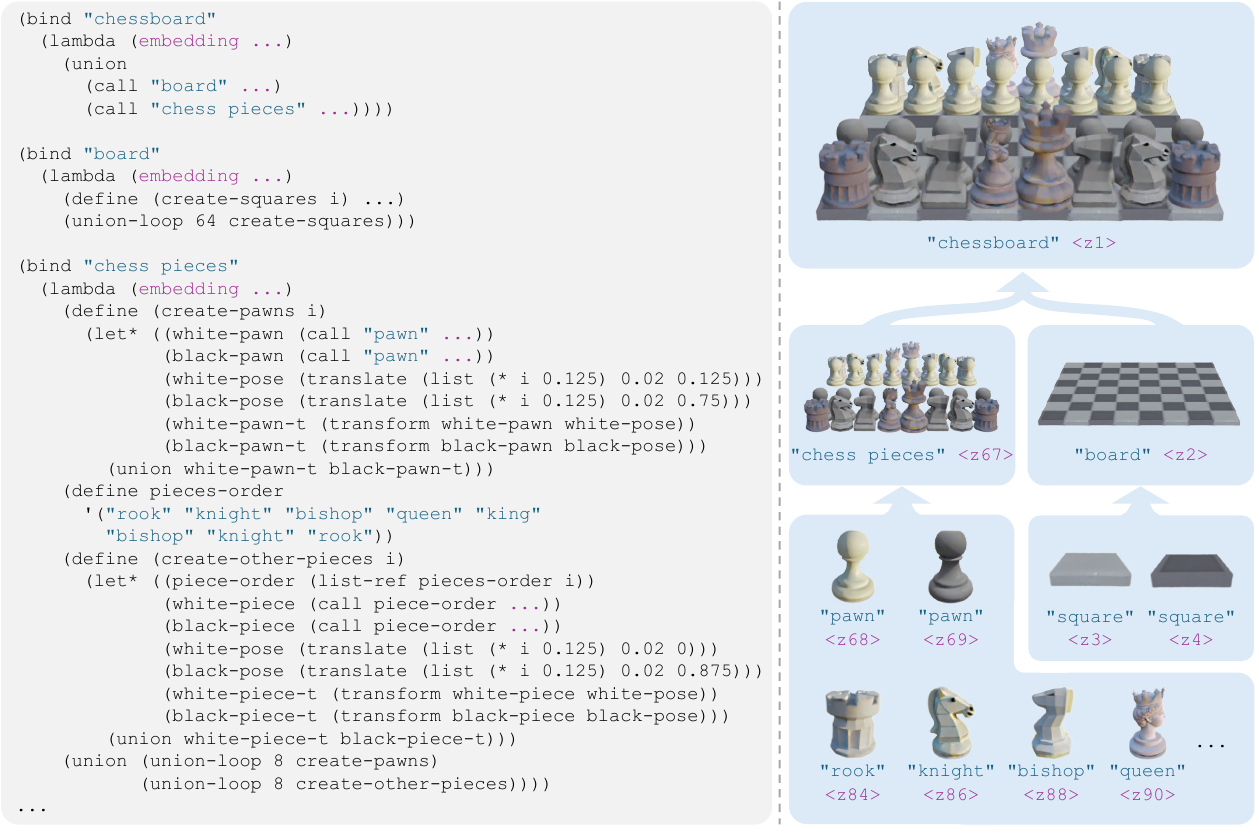}
    \vspace{-2em}
    \caption{\textbf{Overview.}
    A \reprFull represents a scene with three components: a \prog~consisting of entity functions, a set of \words (\eg, \texttt{\textcolor{myWordColor}{pawn}}) denoting the semantic class of the entity functions, and a list of \embds (\eg, \texttt{\textcolor{myEmbdColor}{<z1>}}) capturing the identity of each entity in the scene.
    Each entity function is bound with an entity class name given by a \word, and maps an input \embd to an output entity of that class.
    Executing the \prog evaluates entity functions to compute the full set of entities in the scene. The resulting computation graph, illustrated on the right, captures the dependency structure among entities (indicated by arrows).
    The program shown is converted from our text-conditioned inference method's output, with details included in \cref{app:lm-output-text-cond}; it is written in Lisp-style syntax for brevity and is implemented with Python in practice (\cref{ssec:dsl}).
    }
    \vspace{-1.5em}
    \label{fig:representation}
\end{figure*}
\section{Introduction}
How do you describe a scene? Imagine that you just traveled to Easter Island and would like to explain to Alice the wondrous scene of Ahu Akivi: ``There are seven moai in a row, facing the same direction.'' ``What is a moai?'' Alice asked. ``A moai is a stone human figure without legs, but each also looks slightly different." At this point, you realize it seems difficult to precisely explain the scene using natural language alone. 

In fact, this example highlights a complete scene representation requires at least three types of complementary information:
\emph{structural knowledge}, which is about the joint distribution of multiple instances, like ``seven moai in a row, facing the same direction,'' most naturally described as programs;
\emph{category-level semantics}, which may be shared across instances, often described in words, such as ``moai'';
and \emph{instance-level intrinsics}, tied to the identity of each specific object or part, such as its geometry, color, and texture, which is hard to describe but easy to recognize.

Modern AI techniques provide natural grounding for each of the three modalities, while also falling short of capturing all: 
in-context learning of pre-trained language models (LMs) enables the inference of domain-specific \progs~\citep{brown2020language}; 
LMs themselves capture rich semantic information based on \words in natural language;
and \embds obtained via techniques like textual inversion~\citep{gal2022textual} or low-rank adaptation~\citep{hu2021lora} best capture object identity.   
However, none of these existing representations alone is sufficient for scene generation and editing.

We introduce the \reprFull, a representation that integrates the three modalities---\progs, \words, and \embds---to precisely and concisely describe the structure, semantics, and identity of visual scenes. 
In the \reprFull, a \prog specifies a computation process that defines the organization of a collection of \emph{entities} in the scene, including extrinsics like poses and structural regularity like repetitions. Each entity is associated with a \word referring to its semantic group, as well as an \embd describing its instance-specific attributes.

In addition to the representation itself, we propose a training-free inference module using a pre-trained LM as a backbone to infer the \reprFull from texts and images. 
When provided with a domain-specific language (DSL) for scenes, LMs decompose the task of complex scene generation into simpler tasks of scene component generation by predicting their corresponding modular functions.
We also discuss possible neural, traditional, and hybrid graphics engines that render images based on the representation. Together, the \reprFull, the inference module, and the renderer lead to a system for high-quality, detailed 3D and 4D scene generation and editing. 

In summary, our contributions include
\begin{enumerate}
    \item A scene representation, the \reprFull, capturing structure, semantics, and identity of visual scenes using \progs, \words, and \embds.
    \item A training-free method that infers the representation from texts and/or images using pre-trained LMs.
    \item A generic rendering module that renders the \reprFull into an image.
    \item Empirical results on text- and image-prompted scene generation and editing tasks.
\end{enumerate}
\section{Related Work}

Visual scene representations are arguably the most fundamental problem in computer vision; thus, for sure, we may not enumerate all related work. As our \reprFull comprises \progs, \words, and \embds, we organize our discussion accordingly into three categories: 
scene representations using program-based representations (\cref{ssec:related-programs}), semantic graph-based representations (\cref{ssec:related-semantics}), and a pre-trained generative model's latent space (\cref{ssec:related-latent}). 

\subsection{Representing Scenes as Programs}
\label{ssec:related-programs}

Programs can specify not only the relations among scene components mentioned in \cref{ssec:related-semantics}, but also structural patterns such as hierarchy and repetitions, 
making them suitable as explicit descriptions of scene structures. 
Prior works have proposed to use programs in the form of sequences of execution commands as object-centric representations, followed by neural executors that render the programs into 3D shapes~\citep{tian2018learning,sharma2018csgnet,deng2022unsupervised}. In comparison, ShapeAssembly~\citep{jones2020shapeAssembly} introduces higher-level functions with semantically meaningful function names, \eg, ``chair'' and ``back'', to its program representation. Both ShapeAssembly and ours adopt the design principle of function abstraction, which results in clearly stated hierarchy relation among components and better program editability. However, ShapeAssembly uses cuboids as the shape representation and does not model appearance, while ours allows for more precise geometry and appearance modeling using expressive neural embeddings. 

All representations above require 3D datasets for training. More recently, with the advance of language models (LMs), several methods~\citep{zhou2024scenex,hu2024scenecraft,yamada2024l3go,sun20233d,zhang2023creative,tam2024scenemotifcoder} have proposed to use zero-shot LM inference for generating programs that will be rendered into scenes. These methods operate on top of program syntax from specific graphics renderers such as Blender~\citep{blender}, and they do not permit parameters in high-dimensional embedding spaces unlike ours.

\subsection{Representing Scenes with Semantic Graphs}
\label{ssec:related-semantics}

Prior semantic scene representations often adopt a graph to encode semantic scene components, such as objects and parts. 
In particular, \cite{yuille2006vision,huang2018holistic} propose to employ a parse graph of context-free grammar, using terminal nodes to correspond to objects and their attributes, to represent a scene. Both works employ an analysis-by-synthesis approach to infer the representation from images that heavily rely on domain-specific priors. 
Alternative representations include scene graph~\citep{johnson2015image,johnson2018image,gao2024graphdreamer}, where each node in a graph corresponds to an object and an edge corresponds to a pairwise relation, and StructureNet~\citep{mo2019structurenet}, which focuses on an object-centric setting and uses nodes for object parts. While these representations preserve the high-level semantics of scenes or objects, they leave out low-level precision; thus, geometric, textural, or relational details that cannot be fully specified by language or hand-crafted rules are often ignored. We address this issue via the inclusion of embeddings. 

\subsection{Representing Scenes with Generative Model Latents}
\label{ssec:related-latent}
The latent space of visual generative models can serve as a representation space for visual scenes. Such latent space can effectively capture the exact visual content of scenes, including geometry and appearance details, and can be either directly inferred, \eg, in variational inference~\citep{kingma2013auto} and model inversion~\citep{zhu2016generative}. More recently, text-to-image diffusion models have shown remarkable results in image synthesis. This class of models offers several candidate representation spaces including the space of textual embeddings~\citep{gal2022textual}, low-rank network weights~\citep{hu2021lora}, full model weights~\citep{ruiz2023dreambooth},
or noise vectors in the diffusion process~\citep{song2020denoising,mokady2023null,ho2020denoising}.
However, such representations typically do not offer interpretable semantics or explicitly encode hierarchical scene structures unlike ours. 

\section{The Scene Language}

We aim to design a visual scene representation that encodes the structure, semantics, and visual content of scenes.
Towards this goal, we propose the \reprFull, which represents a scene with three components: a \progVar that encodes scene structure by specifying the existence and relations of scene components, which we will refer to as entities; \words in natural language that denote the semantic group of each entity in the scene; and neural \embds that pertain the low-level visual details and identities of the entities by permitting an expressive input parameter space.
In the following, we will first give a formal definition of the representation (\cref{ssec:formal}), and then introduce a domain-specific language (DSL) (\cref{ssec:dsl}) as its realization.

\begin{table}[t]
\centering
\scriptsize
\setlength{\tabcolsep}{2pt}
\begin{tabular}{lll}
\toprule
\cref{ssec:formal} & \cref{ssec:dsl} & Definition
\\
\midrule
\underline{\textit{Expressions}} \\[0.5em]
$\Psi_\text{transform}$ & \texttt{transform} & Transform an entity\\
$\Psi_\text{union}$ & \texttt{union} & Compose entities \\
$f_w: z, \gamma \mapsto h$ & \texttt{<entity-func>} & Mapping \embds to entity\\
$f_w(z, \gamma)$ & \makecell[l]{\texttt{(call <\word\unskip>} \\\texttt{\quad<\embd\unskip>*)}} & Entity function application \\
\midrule
\underline{\textit{Data Types}} \\[0.5em]
$w$ &\texttt{\WordType} & Describing semantics \\
$t$ &\texttt{Matrix} & Entity pose \\
$z$ &\texttt{\EmbdType} & Specifying entity identity \\
$\gamma$ &\texttt{List[\EmbdType\unskip]} & Specifying descendent entities' identities \\
$h$ & \texttt{Entity} & An entity\\
$s$ & \texttt{Entity} & The represented scene \\
\bottomrule
\end{tabular}
\vspace{-1em}
\caption{\textbf{Summary of Notations in \cref{ssec:formal,ssec:dsl}.}
}
\vspace{-2em}
\label{tab:notations}
\end{table}

\begin{figure*}[t]
    \centering
    \includegraphics[width=\textwidth]{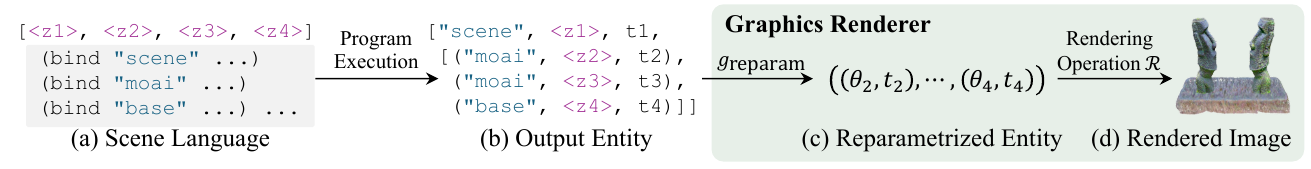}
    \vspace{-2.2em}
    \caption{\textbf{Rendering.} 
    Given a \reprFull in (a), a program interpreter executes the program to obtain a data object in (b). 
    A graphics renderer first reparameterizes the data object from (b) into the renderer-specific parameter space, and then executes the rendering operation $\mathcal{R}$ to obtain the final image output in (d). }
    \vspace{-1.3em}
    \label{fig:rendering}
\end{figure*}
\begin{table}[t]
\centering
\scriptsize
\setlength{\tabcolsep}{1pt}
\begin{tabular}{lccc}
\toprule
 \multirow{2}{*}{\textbf{Renderer}} & \multicolumn{3}{c}{\textbf{Examples}} 
 \\
 \cmidrule(lr){2-4}
& 
Rendering Operation $\mathcal{R}$
& 
Parameters from $\Theta$
&
$g_\text{reparam}$
\\
\midrule
\makecell{Primitive-based} & 
\makecell{Light transport} & 
\makecell{Shape and BRDF} & 
LM inference
\\
Asset-based & 
Ray tracing &
Asset metadata & 
LM inference
\\
SDS-based & 
Gaussian splatting & 
Gaussian parameters & 
SDS optimization
\\
T2I model & 
Feed-forward pass &
Text embeddings in $\mathcal{Z}$ &
CLIP encoding
\\
\bottomrule
\end{tabular}
\vspace{-1em}
\caption{\textbf{Graphics Renderers Examples.} 
}
\vspace{-2em}
\label{tab:renderers}
\end{table}

\subsection{Formal Definition}
\label{ssec:formal}
The \reprFull for a scene $s$, denoted as $\Phi(s)$, is formally defined as follows:
\begin{equation}
   \label{eqn:repr}
   \Phi(s) := \left(W, P, Z\right).
\end{equation}

Here, $W$ is a collection of phrases in natural language, referred to as \words, \eg, $W = \{\text{\textcolor{myWordColor}{board}}, \text{\textcolor{myWordColor}{pawn}}, \cdots\}$ as illustrated in \cref{fig:representation}, and $P := \{f_w\}_{w\in W}$ is a \progVar consisting of a set of entity functions $f_w$ indexed by $w\in W$. Each entity function $f_w$ defines a class of entities in the scene; it is uniquely identified by the associated $w\in W$, which concisely summarizes the semantic meaning of the defined class. 
Each of $f_w$ maps neural \embds $(z, \gamma)$ to a specific entity $h$ in the scene, where $z$ specifies the attributes and identity of $h$, 
like a specific color of a \textcolor{myWordColor}{pawn}, and $\gamma$ specifies identities of its sub-entities.
Hence, the complete \reprFull $\Phi(s)$ of a particular scene $s$ also contains an ordered set of neural \embds $Z := \left[z_1, z_2, \cdots, z_J\right]$ corresponding to the $J$ specific entities $\left[h_1, h_2, \cdots, h_J\right]$ from scene $s$, where $h_j\in J$ is computed by applying an entity function on inputs $z_j\in Z, \gamma_j \subset Z$. 

Crucially, \prog $P$ captures scene structures in three aspects.
First, each entity function $f_w\in P$ transforms and composes multiple sub-entities (\eg, $64$ squares) into a new, more complex entity (\eg, a board), naturally reflecting the hierarchical, part-whole relations in the scene, as the arrows in \cref{fig:representation} highlight.
Second, multiple entities $h_j$'s may belong to the same semantic class $w$ (\eg, \textcolor{myWordColor}{square}), and can thus be represented by reusing the same entity function $f_w$ with distinct \embds $z_j$'s.
Finally, each entity function also captures the precise spatial layout of the sub-entities by specifying their relative poses during the composition, such as $64$ squares forming an $8 \times 8$ grid.

The following expands on how functions from $P$ are defined and the program execution procedure to compute the represented scene $s$.
Notations are summarized in \cref{tab:notations}.

\paragraph{Entity Function Definitions.}
An entity function $f_w$ outputs an entity $h$ of semantic class $w$. The function $f_w$ takes two inputs: an \embd $z$ that specifies $h$'s identity, as well as an ordered set $\gamma$ containing \embds of all descendent entities of $h$. Denote $\{h^{(i)}\}_{i=1}^N$ as the $N$ sub-entities of $h$, where $N$ is a constant determined by $f_w$. Then we have $\gamma = \left\{z^{(i)}, \gamma^{(i)}\right\}_{i=1}^N$, where $z^{(i)}$ and $\gamma^{(i)}$ are the embeddings of $h^{(i)}$ and of the descendent entities of $h^{(i)}$, respectively. Thus, $f_w(z, \gamma)$ is recursively defined as
\begin{equation}
    h = f_w(z, \gamma):= \Psi_\text{union}\left(\left\{\Psi_\text{transform}(h^{(i)}, t^{(i)}(z))\right\}_{i=1}^N\right), \label{eqn:Psi-f}
\end{equation}
where $h^{(i)} = f_{w^{(i)}}(z^{(i)}, \gamma^{(i)})$ is a sub-entity of $h$ and $t^{(i)}(z)$ specifies its pose.

Here, $f_w$ specifies the computation procedure to obtain output $h$ from $N$ sub-entities $\{h^{(i)}\}_{i=1}^N$ via two operations: $\Psi_\text{transform}$ applies an input-dependent pose $t^{(i)}(z)$ to a sub-entity $h^{(i)}$, transforming it from its canonical frame to the world frame of $h$, and $\Psi_\text{union}$ composes multiple sub-entities into one single entity.
Each sub-entity $h^{(i)}$ is computed by recursively applying an entity function $f_{w^{(i)}}$ also defined using \cref{eqn:Psi-f}.
For instance, let $f_w$ denote the entity function that produces the board in \cref{fig:representation} (namely, $w = \text{\textcolor{myWordColor}{board}}$).
This function $f_w$ composes $64$ sub-entities $\{h^{(i)}\}_{i=1}^{64}$ of the same class \textcolor{myWordColor}{square}, which are in turn obtained by executing the \emph{same} entity function $f_{w^{(i)}} = f_\text{\textcolor{myWordColor}{square}}$ with \emph{different} \embds $z^{(i)}$ and $\gamma^{(i)}$.

\paragraph{Program Execution.}
To obtain a scene $s$ from the \reprFull $\Phi(s) = (W, P, Z)$, a program executor identifies a root entity function $f_{w_1}$ from $P$ that is not dependent by any other function (\eg, $w_1 = \text{\textcolor{myWordColor}{chessboard}}$ from \cref{fig:representation}), and evaluates this root function using the first \embd $z_1 \in Z$ and the rest as its descendants, $\gamma_1 := [z_2, \cdots, z_J]\subset Z$, to obtain $s = f_{w_1}(z_1, \gamma_1)$.
Evaluating $f_{w_1}(z_1, \gamma_1)$ expands the computation recursively to its children functions $h_j = f_{w_j}(z_j, \gamma_j)$ as defined in \cref{eqn:Psi-f}, obtaining a full sequence of all the entities $h_j$ of the scene, where $j=1, 2, \cdots, J, z_j \in Z, w_j\in W$. The order in $Z$ corresponds to the depth-first-search of the computation graph starting from $z_1$. An example computation graph is visualized on the right of \cref{fig:representation}. 

\subsection{Scene Language as Programming Language}
\label{ssec:dsl}
We now concretize the definition in \cref{ssec:formal} with a domain-specific language (DSL) specified in \cref{app:dsl4d}.
We introduce macro operations \texttt{union} for $\Psi_\text{union}$, \texttt{union-loop} which calls \texttt{union} on entities evaluated in a for-loop, and \texttt{transform} for $\Psi_\text{transform}$. 
Together with \texttt{call} which applies function calls, these four macro operations fully define entity functions in the DSL. 
Entity functions are identified with the associated \words in the DSL via two special forms: \texttt{bind}, which binds entity function $f_w$ to \word $w$, and \texttt{retrieve}, which retrieves $f_w$ given $w$ and applies $f_w$ on actual \embd parameters. If $w$ is never bound to a function, it corresponds to an entity function with no sub-entities (\ie, $N=0$ in \cref{eqn:Psi-f}), in which case \texttt{call} returns a primitive entity with no children. 

The data type of an entity $h$ (\cref{eqn:Psi-f}) is denoted as \texttt{Entity}. It stores two data fields \texttt{\WordType} and \texttt{\EmbdType}, describing the entity's semantic group and identity respectively, and stores each children entity together with its pose in the frame of $h$.
In particular, \texttt{\EmbdType} captures the visual details and requires a highly expressive representation, \eg, neural embeddings. We employ the textual embedding space of OpenCLIP-ViT/H~\citep{ilharco_gabriel_2021_5143773} for parameterization, denoted as $\mathcal{Z}_\text{CLIP}$. It offers the advantage that embeddings can be either encoded directly from natural language or inferred from images with Textual Inversion~\citep{gal2022textual}.
\cref{tab:notations} summarizes the expressions and data types in accordance with the notations from \cref{ssec:formal}.

\section{Rendering}
\label{ssec:render}

Applying the proposed scene representation to image generation tasks requires rendering a \reprFull $\Phi(s)$ into images.
To do so, first, the program interpreter evaluates $\Phi(s)$ to obtain a data object of type \texttt{Entity}. 
Afterward, a graphics renderer maps the \texttt{Entity} data object to its rendering parameter space and renders it into a final image.

\paragraph{Renderer Specifications.}
We define the specification of a graphics renderer, a module in the proposed representation, as follows. A graphics renderer is determined by 
primitive parameter space $\Theta$
and a rendering operation $\mathcal{R}: \mathcal{P}\left(\Theta\times \mathcal{T}\right)\rightarrow \mathcal{I}$,
where $\mathcal{T}$ is the space of 3D affine transformations representing poses, $\mathcal{P}$ denotes all possible subsets, and $\mathcal{I}$ is the space of rendered images. 
In order to determine a mapping from program execution outputs of type \texttt{Entity} (\cref{fig:rendering}b) to the admissible input domain of rendering operation $\mathcal{R}$ (\cref{fig:rendering}c), we assume access to a reparameterization function $g_\text{reparam}$ that maps from \texttt{Tuple[\WordType, \EmbdType\unskip]} to $\Theta$, 
and compute a primitive's pose $t\in\mathcal{T}$ by multiplying all $\texttt{Matrix}$ values along the path from the primitive to the scene (root) in the entity hierarchy, analogous to computing limb poses in a kinematic tree.

\paragraph{Renderer Instantiations.}
An example renderer instantiation is with
Score Distillation Sampling (SDS)~\citep{poole2022dreamfusion} guidance, where $\Theta$ is a differentiable 3D representation, and we specify $g_\text{reparam}$ as follows. 
Recall that an entity is associated with a \texttt{\Word} value, \eg, \texttt{\textcolor{myWordColor}{moai}} for one statue from \cref{fig:teaser,fig:rendering}, and an \texttt{\Embd} value, \eg, \texttt{\textcolor{myEmbdColor}{<z2>}}. 
For each primitive entity (\ie, an entity with no children), given these two value fields together with the fields of its ancestors, we use a manually specified language template $c$, or \texttt{a \textcolor{myEmbdColor}{<z2>} \textcolor{myWordColor}{moai}, in the style of \textcolor{myEmbdColor}{<z1>}, 3D model} in this example, to embed them into $z = g_\text{CLIP}(c) \in\mathcal{Z}_\text{CLIP}$ where $g_\text{CLIP}$ is the pre-trained CLIP text encoder. 
Then, $g_\text{reparam}: \mathcal{Z}_\text{CLIP}\rightarrow\Theta$ corresponds to the SDS-guided optimization to find a solution in $\Theta$ that aligns with the input condition $z\in\mathcal{Z}_\text{CLIP}$. Output 3D scenes can be personalized by editing embeddings, \eg, \texttt{\textcolor{myEmbdColor}{<z1>}} which controls the global style in \cref{fig:teaser}. 

For the underlying 3D representation, we use 3D Gaussian Splatting~\citep{kerbl3Dgaussians}  where images are rendered by splatting a set of 3D Gaussians onto the image plane; other differentiable 3D representations such as neural fields will also be suitable. 
We base our implementation on GALA3D~\citep{zhou2024gala3d}, and use MVDream~\citep{shi2023mvdream} and a depth-conditioned ControlNet~\citep{zhang2023controlnet} for guidance.

We refer to the renderer above as the Gaussians renderer. Other possible renderers include primitive-based renderers, such as Mitsuba~\citep{Mitsuba3} with graphics primitives of cubes, spheres, and cylinders; asset-based game engines, such as MineCraft~\citep{minecraft}; and feed-forward inference of layout-conditioned text-to-image (T2I) diffusion models, such as MIGC~\citep{zhou2024migc}, which achieves 2D bounding box conditioning by controlling attention layers from Stable Diffusion~\citep{rombach2022high}). \cref{tab:renderers} shows a summary; details are in \cref{app:renderers}. 

\section{Inference via Pre-Trained Language Models}
\label{sec:inference}

We introduce a training-free method to infer the representation $\Phi(s) = (W, P, Z)$ from text or image descriptions of a scene $s$. As explained below, we first prompt a pre-trained language model (LM) to generate the non-neural components ($W, P$) and then obtain neural embeddings ($Z$) from texts via the CLIP text encoder or from images with a pre-trained text-to-image diffusion model.

LMs have shown remarkable capability in code generation with common programming languages such as Python. In our implementation, we prompt LMs to generate Python scripts.
We prompt the LM with the input condition, \ie, a scene description in texts or an image; a Python script of helper functions converted from the DSL in \cref{ssec:dsl}; and an example script using the helper functions. 
We use Claude 3.5 Sonnet~\citep{claude3} for all experiments for our method and LM-dependent baselines. 
Full LM prompts are in \cref{app:lm-prompt}.
Function arguments in the LM-generated scripts, which are numeric values or string tokens, are converted to embeddings from $\mathcal{Z}_\text{CLIP}$ (\cref{ssec:dsl}) using language templates and the CLIP text encoder $g_\text{CLIP}$. 
For example, in the raw LM output, function calls for white pieces in \cref{fig:representation} have input attribute \texttt{\{"color":(.9,.9,.9)\}}, and we prompt LM to describe the color value as a word, and feed the word into $g_\text{CLIP}$ to compute \texttt{<z68>}.
For image-conditioned tasks, for each primitive entity in the execution output of $P$, we first use GroundingSAM~\citep{kirillov2023segment,ren2024grounded} to segment out the region defined by the word associated with the entity. We then use Textual Inversion~\citep{gal2022textual} to optimize an embedding to reconstruct the cropped image with the diffusion model training objective. The full process is deferred to \cref{app:textual-inversion}.
\begin{figure}[t]
    \centering
    \includegraphics[trim={0pt 0pt 0pt 0pt}, clip, width=0.95\linewidth]{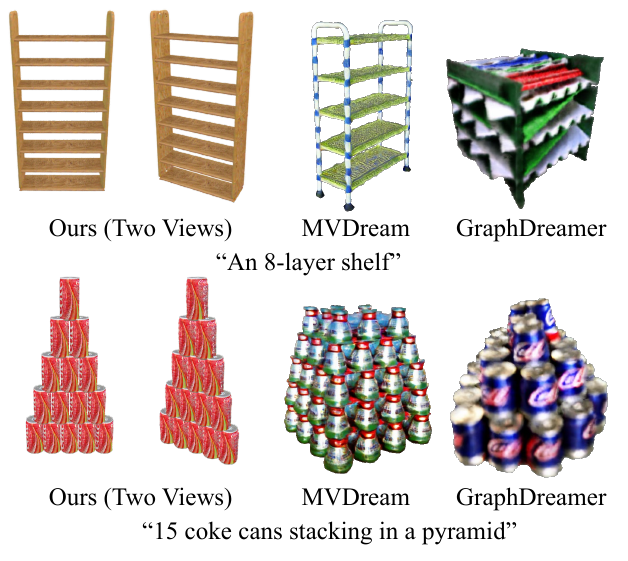}
    \vspace{-1.2em}
    \caption{\textbf{Text-Prompted Scene Generation.} 
    Ours produce more accurate outputs than using no intermediate representation~(MVDream) or scene graph~(GraphDreamer). 
    }
    \label{fig:generation}
    \vspace{-0.4em}
\end{figure}

\begin{table}[t]
\centering
\tiny
\setlength{\tabcolsep}{2pt}
\begin{tabular}{lccccc}
\toprule
 & \multicolumn{3}{c}{Numeric Prompts} & \multicolumn{2}{c}{Generic Prompts} \\
\cmidrule(lr){2-4} \cmidrule(lr){5-6}
Methods & Alignment ($\uparrow$) & CLIP ($\uparrow$) & Count ($\uparrow$) & Alignment ($\uparrow$) & CLIP ($\uparrow$) \\
\midrule
GraphDreamer~\citep{gao2024graphdreamer} 
& $3.56$\xpm{7.38} & $0.297$\xpm{0.085} & 0.11 
& $15.75$\xpm{20.39} & $0.230$\xpm{0.097} \\
MVDream~\citep{shi2023mvdream} 
& $10.79$\xpm{12.83} & $0.312$\xpm{0.046} & 0.11 
& $35.13$\xpm{31.90} & $\mathbf{0.328}$\xpm{\mathbf{0.031}} \\
Ours 
& $\mathbf{85.65}$\xpm{\textbf{13.71}} & $\mathbf{0.351}$\xpm{\textbf{0.051}} & $\mathbf{1.0}$ 
& $\mathbf{52.13}$\xpm{\mathbf{26.83}} & $\mathbf{0.328}$\xpm{\mathbf{0.030}} \\
\bottomrule
\end{tabular}
\vspace{-1.5em}
\caption{\textbf{Evaluation} for text-prompted 3D generation. We report the following metrics averaged across $9$ numeric and $8$ generic prompts: the percentages of user preferences for prompt alignment, CLIP similarity scores, and counting accuracy (numeric prompts only, $0$ for inaccurate and $1$ for accurate). Here $\pm$ denotes standard deviation across scenes.}
\vspace{-2.4em}
\label{tab:user-study}
\end{table}

\begin{figure}
    \centering
    \includegraphics[width=\linewidth]{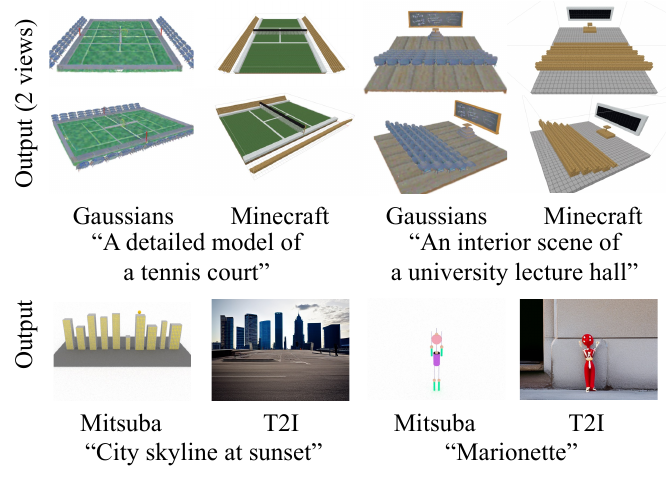}
    \vspace{-2.4em}
    \caption{\textbf{Renderings Across Graphics Renderers.}
    Different renderers produce renderings that adhere to the same representation and are therefore visually aligned, while each exhibits a different imaging style. 
    Text inputs are shown at the subfigure bottoms. 
    \vspace{-1.5em}
    }
    \label{fig:aligned}
\end{figure}

\begin{figure*}[t]
        \centering
        \includegraphics[width=\linewidth]{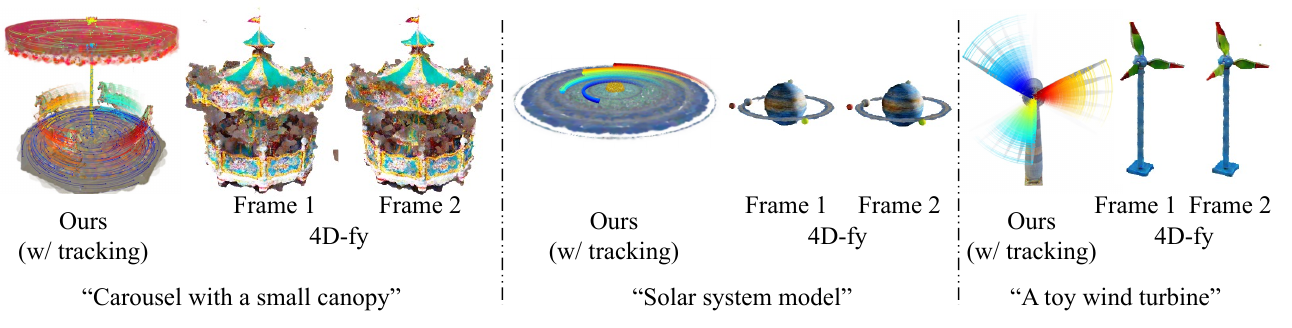}
        \vspace{-2.4em}
        \caption{\textbf{Text-Prompted 4D Scene Generation.} The proposed representation may also capture scene dynamics and be applied for synthesizing 4D scenes. It explicitly represents the temporal correspondence of an entity in a dynamic scene. Each colored trajectory denotes tracking of a temporally moving point. 
        Compared to 4D-fy, our method produces more salient motion. 
        }
        \vspace{-.6em}
        \label{fig:generation4d}
\end{figure*}

\begin{figure*}[t]
    \centering
    \includegraphics[width=\linewidth]{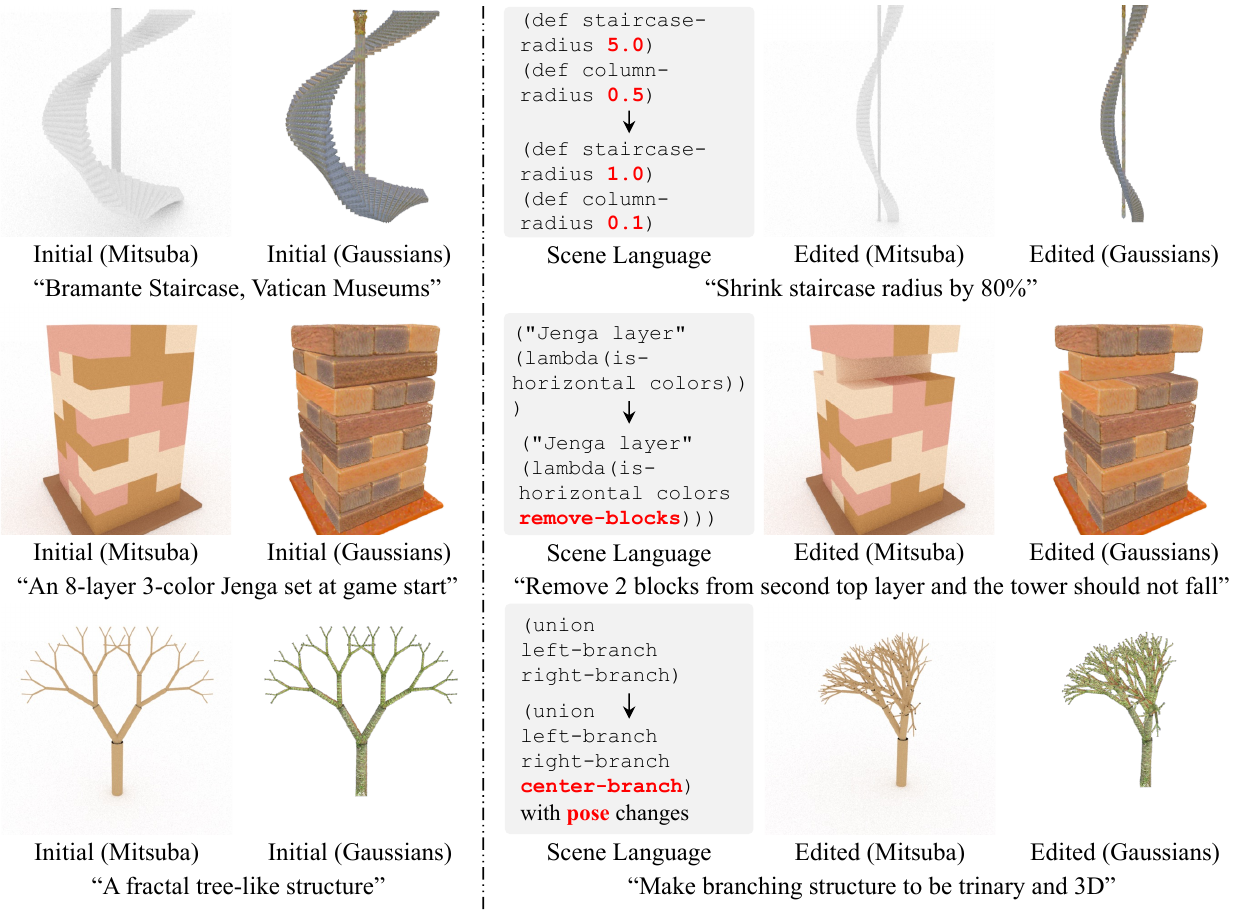}
    \vspace{-2.2em}
    \caption{\textbf{Scene Editing with Language Instructions.} 
    Our representation is editing-friendly. 
    Bottoms of each row show initial scene descriptions and editing instructions in the format of user text prompts. 
    We prompt an LM to infer the initial \reprFull as well as the edits (shown with texts in red), and show image renderings with two renderers. 
    }
    \vspace{-1.6em}
    \label{fig:editing}
\end{figure*}

\begin{figure*}[t]
    \centering
    \includegraphics[trim={0pt 0pt 40pt 30pt}, clip, width=\linewidth]{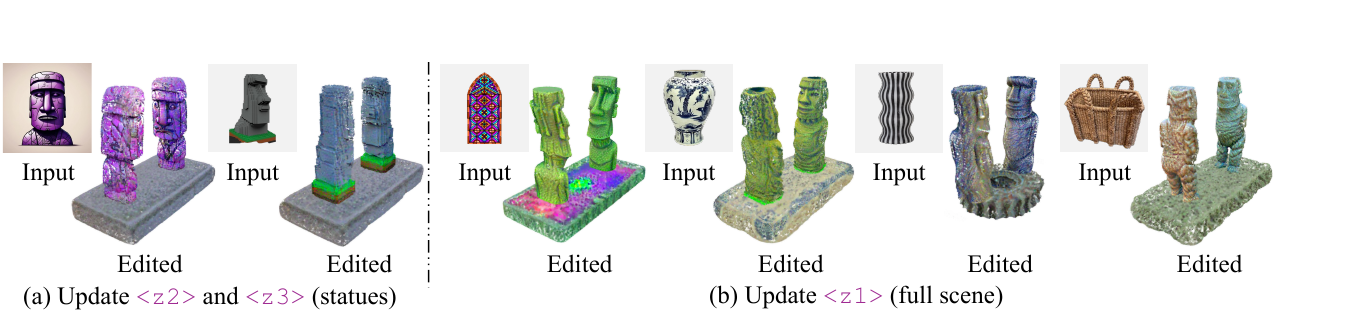}
    \vspace{-2.5em}
    \caption{\textbf{Scene Editing with Image Instructions.} 
    This figure shows style transfer results under the same setup as in \cref{fig:teaser}. Replacing embeddings of target entities in a \reprFull with a new embedding specified by a user image achieves the editing effect.
    }
    \label{fig:style_transfer}
    \vspace{-1.4em}
\end{figure*}

\begin{figure}[t]
    \centering
    \includegraphics[trim={0pt 0pt 5pt 0pt}, clip, width=\linewidth]{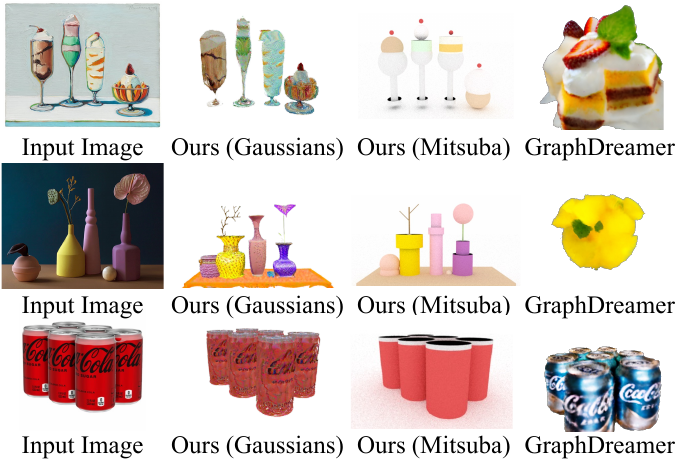}
    \vspace{-2.4em}
    \caption{\textbf{Image-Prompted Scene Generation.} 
    Compared to scene graphs from GraphDreamer, programs from our representation encode additional scene structure, \eg, repetitions, and specify geometric entity relations more precisely.
    Embeddings from ours further enable visual identity preservation in the outputs.
    }
    \label{fig:reconstruction}
    \vspace{-1.8em}
\end{figure}

\section{Experiments}

\subsection{Text-Prompted Generation and Editing}
\label{ssec:text-cond-generation}
\paragraph{Baselines.}
To evaluate our representation in text-prompted 3D generation task, we compare our inference pipeline with 3D scene generation methods using alternative intermediate representations, \eg, scene graph. In particular, we compare with GraphDreamer~\citep{gao2024graphdreamer} as an exemplar approach, which generates scene graphs from texts via LM prompting and then synthesizes scenes conditioned on the graphs via SDS guidance. 
We further ablate the role of structural representation in this task by comparing
ours with the backbone of our SDS-based renderer, MVDream~\citep{shi2023mvdream}, as a direct scene generation approach. 
Details are in \cref{app:graphdreamer}. 

\begin{table}[t]
\scriptsize
\setlength{\tabcolsep}{2pt}
    \begin{minipage}{0.55\linewidth}
        \centering
        \begin{tabular}{lcc}
            \toprule
            Methods & 
            CLIP~\citep{ilharco_gabriel_2021_5143773} ($\uparrow$)
            & Dynamic~\citep{huang2024vbench} ($\uparrow$) \\ \midrule
            4D-fy~\citep{bahmani20244dfy} & \textbf{0.352} & $0.2\%$ \\
            Ours & 0.341 & $\textbf{5.9\%}$ \\ \bottomrule
        \end{tabular}
    \end{minipage}
    \begin{minipage}{0.44\linewidth}
        \centering
        \begin{tabular}{lc}
            \toprule
            Methods & LPIPS~\citep{zhang2018unreasonable} ($\downarrow$) \\ \midrule
            GraphDreamer~\citep{gao2024graphdreamer} & 0.811 \\
            Ours & \textbf{0.681} \\ \bottomrule
        \end{tabular}
    \end{minipage}
    \vspace{-1em}
    \caption{\textbf{Evaluation.} Left: For text-prompted 4D generation, 4D-fy achieves higher prompt alignment measured with CLIP similarity but has small motion. 
    Right: For image-prompted scene generation, our results are better aligned with input image prompts compared to GraphDreamer. 
    }
    \label{tab:lpips-flow}
    \vspace{-2em}
\end{table}

\paragraph{Metrics.}
We conduct a user study where $103$ users are asked to choose one of the three videos, showing 360-renderings from ours and two baselines in a randomized order, that aligns the best with text prompts. Following \cite{gao2024graphdreamer}, for each scene, we compute the cosine similarity of CLIP~\citep{radford2021learning,ilharco_gabriel_2021_5143773} embeddings between a rendering and the text prompt and report the average over $120$ viewpoints.
We further report whether outputs have the correct object counts. Details are deferred to \cref{app:metrics}.

\paragraph{Results.}
\cref{tab:user-study} shows that our method achieves a more favorable prompt alignment than the baselines as measured in the user study and CLIP similarity, and has a clear advantage in counting accuracy. 

\cref{fig:generation} and \cref{app:exp-generation} show qualitative results of ours using the SDS-based renderer. Compared to the direct 3D scene generation method MVDream, our approach is compositional and adheres more closely to input prompts in scenes involving multiple objects. 
Compared to a scene graph representation, where entity relations are restricted to be between two objects and are bottlenecked by the coarseness of natural language descriptions, \eg, ``aligned in a row'', a program-based representation offers more flexible and precise specifications for relations, \eg, the particular coke can arrangement in \cref{fig:generation}. 
This brings the practical benefit of offloading the burden of generating scenes involving complex entity relations from the T2I model (used for SDS guidance in both ours and GraphDreamer) towards LM, leading to accurate and detailed generation results.

\paragraph{Across Graphics Renderers.}
\cref{fig:aligned} shows that our representation can be rendered with different renderers (\cref{ssec:render}), producing aligned outputs in different visual domains.

\paragraph{4D Generation.}
The inference method from \cref{sec:inference} can be applied to 4D scene generation.
The 4D scene representation is identical to \cref{eqn:repr}, except that there is an additional 4D entity function in the program $P$.

We compare with 4D-fy~\cite{bahmani20244dfy} as an exemplar text-to-4D generation method. 
We measure the prompt alignment of output dynamic renderings using CLIP similarity between rendered frames and input texts, and the motion magnitude using dynamic degrees~\citep{huang2024vbench} computed using RAFT~\citep{teed2020raft} optical flow estimation over neighboring frames. 
4D-fy results have a higher per-frame alignment score than ours but present small dynamic motions.
See \cref{fig:generation4d}, the webpage, and \cref{app:exp-4d} for further results. 

\paragraph{Scene Editing.}
Scenes synthesized from our proposed representation can further be edited by prompting LM with its previously generated script and an editing instruction in natural language instruction. 
Results are shown in \cref{fig:editing}. 
Our representation provides an interpretable and intuitive interface for scene editing, as functions have explicit semantic meanings associated with words, and function reuse significantly improves program readability. 
Furthermore, since the structure of programs reflects the structure of scenes, editing program parameters leads to precise changes in the scenes while preserving the original structure, \eg, the circular arrangement of staircases in \cref{fig:editing}. Desirable editing effects involving multiple primitives, or all staircases in this example, can be effectively achieved via only minor changes in the program space.
Finally, the program structure itself, such as the function header in the Jenga set example, can be adjusted for editing to achieve localized edits that only affect relevant parts of the scene. Further results are shown in \cref{app:exp-editing}.

\subsection{Image-Prompted Generation and Editing}
\label{ssec:image-cond-generation}

Texts provide an easy user interface for scene generation as shown in \cref{ssec:text-cond-generation}, and images allow users to specify further intentions that could be hard to describe with natural language alone.  
Integrating neural embeddings increases the expressivity of the \reprFull and allows it to encode visual details from image prompts, and as a result, enhances our method's controllability in image-prompted scene generation and editing tasks.

\paragraph{Generation.}
\cref{fig:reconstruction} shows results on the task of generating 3D scenes consistent with the structure and content from a user-provided image, termed as the ``inverse semantics'' task in GraphDreamer~\citep{gao2024graphdreamer}. 
We report the LPIPS~\citep{zhang2018unreasonable} score, \ie, the AlexNet~\citep{krizhevsky2014one} feature dissimilarity, between an input image and an output rendering, averaged over 3 scenes with $20$ viewpoints each in \cref{tab:lpips-flow}. 
Embeddings allow our representation to encode visual details from inputs, such as the Coke can texture in \cref{fig:reconstruction}, which is not captured in scene graphs from GraphDreamer using per-object natural language descriptions. This is reflected quantitatively in \cref{tab:lpips-flow}. 

\paragraph{Editing.}
Our representation applies to image-prompted editing tasks such as style transfer (\cref{fig:teaser}). 
Edits can target partial or full scenes as shown in \cref{fig:style_transfer}(a) and (b), respectively, by updating embeddings for corresponding entities. 

\section{Conclusion}
We have introduced a visual scene representation, termed the \reprFull, which encodes three key aspects of visual scenes: scene structure, such as hierarchy and repetition, specified via programs; semantics of individual components succinctly summarized via words; and identities of each component precisely captured via neural embeddings.
We formalize the representation as a programming language defined using a DSL.
We show that the \reprFull can be efficiently inferred from both text and image inputs leveraging pre-trained language models.
Once the program is executed, the resulting scene can be rendered into images using a variety of graphics renderers.
Our \reprFull produces 3D and 4D scenes with high fidelity and enables easy and precise editing. 

\paragraph{Acknowledgments.}
This work is in part supported by AFOSR YIP FA9550-23-1-0127, ONR N00014-23-1-2355, ONR YIP N00014-24-1-2117, ONR MURI N00014-22-1-2740, and NSF RI \#2211258 and \#2338203. YZ is in part supported by the Stanford Interdisciplinary Graduate Fellowship.

{
    \small
    \bibliographystyle{ieeenat_fullname}
    \bibliography{ICLR-25/main}

\begin{thebibliography}{45}
\providecommand{\natexlab}[1]{#1}
\providecommand{\url}[1]{\texttt{#1}}
\expandafter\ifx\csname urlstyle\endcsname\relax
  \providecommand{\doi}[1]{doi: #1}\else
  \providecommand{\doi}{doi: \begingroup \urlstyle{rm}\Url}\fi

\bibitem[Anthropic(2024)]{claude3}
Anthropic.
\newblock The {Claude} 3 model family: {Opus, Sonnet, Haiku}, 2024.

\bibitem[Bahmani et~al.(2024)Bahmani, Skorokhodov, Rong, Wetzstein, Guibas, Wonka, Tulyakov, Park, Tagliasacchi, and Lindell]{bahmani20244dfy}
Sherwin Bahmani, Ivan Skorokhodov, Victor Rong, Gordon Wetzstein, Leonidas Guibas, Peter Wonka, Sergey Tulyakov, Jeong~Joon Park, Andrea Tagliasacchi, and David~B. Lindell.
\newblock {4D}-fy: Text-to-{4D} generation using hybrid score distillation sampling.
\newblock \emph{CVPR}, 2024.

\bibitem[Brown et~al.(2020)Brown, Mann, Ryder, Subbiah, Kaplan, Dhariwal, Neelakantan, Shyam, Sastry, Askell, et~al.]{brown2020language}
Tom Brown, Benjamin Mann, Nick Ryder, Melanie Subbiah, Jared~D Kaplan, Prafulla Dhariwal, Arvind Neelakantan, Pranav Shyam, Girish Sastry, Amanda Askell, et~al.
\newblock Language models are few-shot learners.
\newblock In \emph{NeurIPS}, 2020.

\bibitem[Community(1994)]{blender}
Blender~Online Community.
\newblock Blender, 1994.

\bibitem[Deng et~al.(2022)Deng, Kulal, Dong, Deng, Tian, and Wu]{deng2022unsupervised}
Boyang Deng, Sumith Kulal, Zhengyang Dong, Congyue Deng, Yonglong Tian, and Jiajun Wu.
\newblock Unsupervised learning of shape programs with repeatable implicit parts.
\newblock In \emph{NeurIPS}, 2022.

\bibitem[Gal et~al.(2023)Gal, Alaluf, Atzmon, Patashnik, Bermano, Chechik, and Cohen-Or]{gal2022textual}
Rinon Gal, Yuval Alaluf, Yuval Atzmon, Or Patashnik, Amit~H. Bermano, Gal Chechik, and Daniel Cohen-Or.
\newblock An image is worth one word: Personalizing text-to-image generation using textual inversion.
\newblock In \emph{ICLR}, 2023.

\bibitem[Gao et~al.(2024)Gao, Liu, Chen, Geiger, and Schölkopf]{gao2024graphdreamer}
Gege Gao, Weiyang Liu, Anpei Chen, Andreas Geiger, and Bernhard Schölkopf.
\newblock {GraphDreamer}: Compositional {3D} scene synthesis from scene graphs.
\newblock In \emph{CVPR}, 2024.

\bibitem[Ho et~al.(2020)Ho, Jain, and Abbeel]{ho2020denoising}
Jonathan Ho, Ajay Jain, and Pieter Abbeel.
\newblock Denoising diffusion probabilistic models.
\newblock In \emph{NeurIPS}, 2020.

\bibitem[Hu et~al.(2022)Hu, Shen, Wallis, Allen-Zhu, Li, Wang, Wang, and Chen]{hu2021lora}
Edward~J Hu, Yelong Shen, Phillip Wallis, Zeyuan Allen-Zhu, Yuanzhi Li, Shean Wang, Lu Wang, and Weizhu Chen.
\newblock {LoRA}: Low-rank adaptation of large language models.
\newblock In \emph{ICLR}, 2022.

\bibitem[Hu et~al.(2024)Hu, Iscen, Jain, Kipf, Yue, Ross, Schmid, and Fathi]{hu2024scenecraft}
Ziniu Hu, Ahmet Iscen, Aashi Jain, Thomas Kipf, Yisong Yue, David~A Ross, Cordelia Schmid, and Alireza Fathi.
\newblock {SceneCraft}: An {LLM} agent for synthesizing {3D} scene as {Blender} code.
\newblock In \emph{ICLR Workshop}, 2024.

\bibitem[Huang et~al.(2018)Huang, Qi, Zhu, Xiao, Xu, and Zhu]{huang2018holistic}
Siyuan Huang, Siyuan Qi, Yixin Zhu, Yinxue Xiao, Yuanlu Xu, and Song-Chun Zhu.
\newblock Holistic {3D} scene parsing and reconstruction from a single {RGB} image.
\newblock In \emph{ECCV}, 2018.

\bibitem[Huang et~al.(2024)Huang, He, Yu, Zhang, Si, Jiang, Zhang, Wu, Jin, Chanpaisit, et~al.]{huang2024vbench}
Ziqi Huang, Yinan He, Jiashuo Yu, Fan Zhang, Chenyang Si, Yuming Jiang, Yuanhan Zhang, Tianxing Wu, Qingyang Jin, Nattapol Chanpaisit, et~al.
\newblock Vbench: Comprehensive benchmark suite for video generative models.
\newblock In \emph{CVPR}, 2024.

\bibitem[Ilharco et~al.(2021)Ilharco, Wortsman, Wightman, Gordon, Carlini, Taori, Dave, Shankar, Namkoong, Miller, Hajishirzi, Farhadi, and Schmidt]{ilharco_gabriel_2021_5143773}
Gabriel Ilharco, Mitchell Wortsman, Ross Wightman, Cade Gordon, Nicholas Carlini, Rohan Taori, Achal Dave, Vaishaal Shankar, Hongseok Namkoong, John Miller, Hannaneh Hajishirzi, Ali Farhadi, and Ludwig Schmidt.
\newblock {OpenCLIP}, 2021.

\bibitem[Jakob et~al.(2022)Jakob, Speierer, Roussel, Nimier-David, Vicini, Zeltner, Nicolet, Crespo, Leroy, and Zhang]{Mitsuba3}
Wenzel Jakob, Sébastien Speierer, Nicolas Roussel, Merlin Nimier-David, Delio Vicini, Tizian Zeltner, Baptiste Nicolet, Miguel Crespo, Vincent Leroy, and Ziyi Zhang.
\newblock Mitsuba 3 renderer, 2022.

\bibitem[Johnson et~al.(2015)Johnson, Krishna, Stark, Li, Shamma, Bernstein, and Fei-Fei]{johnson2015image}
Justin Johnson, Ranjay Krishna, Michael Stark, Li-Jia Li, David Shamma, Michael Bernstein, and Li Fei-Fei.
\newblock Image retrieval using scene graphs.
\newblock In \emph{CVPR}, 2015.

\bibitem[Johnson et~al.(2018)Johnson, Gupta, and Fei-Fei]{johnson2018image}
Justin Johnson, Agrim Gupta, and Li Fei-Fei.
\newblock Image generation from scene graphs.
\newblock In \emph{CVPR}, 2018.

\bibitem[Jones et~al.(2020)Jones, Barton, Xu, Wang, Jiang, Guerrero, Mitra, and Ritchie]{jones2020shapeAssembly}
R.~Kenny Jones, Theresa Barton, Xianghao Xu, Kai Wang, Ellen Jiang, Paul Guerrero, Niloy Mitra, and Daniel Ritchie.
\newblock {ShapeAssembly}: Learning to generate programs for {3D} shape structure synthesis.
\newblock \emph{ACM TOG}, 39\penalty0 (6):\penalty0 1--20, 2020.

\bibitem[Kerbl et~al.(2023)Kerbl, Kopanas, Leimk{\"u}hler, and Drettakis]{kerbl3Dgaussians}
Bernhard Kerbl, Georgios Kopanas, Thomas Leimk{\"u}hler, and George Drettakis.
\newblock {3D} {Gaussian} splatting for real-time radiance field rendering.
\newblock \emph{ACM TOG}, 42\penalty0 (4):\penalty0 1--14, 2023.

\bibitem[Kingma(2014)]{kingma2013auto}
Diederik~P Kingma.
\newblock Auto-encoding variational bayes.
\newblock In \emph{ICLR}, 2014.

\bibitem[Kirillov et~al.(2023)Kirillov, Mintun, Ravi, Mao, Rolland, Gustafson, Xiao, Whitehead, Berg, Lo, et~al.]{kirillov2023segment}
Alexander Kirillov, Eric Mintun, Nikhila Ravi, Hanzi Mao, Chloe Rolland, Laura Gustafson, Tete Xiao, Spencer Whitehead, Alexander~C Berg, Wan-Yen Lo, et~al.
\newblock Segment anything.
\newblock In \emph{ICCV}, 2023.

\bibitem[Krizhevsky(2014)]{krizhevsky2014one}
Alex Krizhevsky.
\newblock One weird trick for parallelizing convolutional neural networks.
\newblock \emph{arXiv preprint arXiv:1404.5997}, 2014.

\bibitem[Mo et~al.(2019)Mo, Guerrero, Yi, Su, Wonka, Mitra, and Guibas]{mo2019structurenet}
Kaichun Mo, Paul Guerrero, Li Yi, Hao Su, Peter Wonka, Niloy Mitra, and Leonidas Guibas.
\newblock {StructureNet}: Hierarchical graph networks for {3D} shape generation.
\newblock \emph{ACM TOG}, 38\penalty0 (6):\penalty0 1--19, 2019.

\bibitem[{Mojang Studios}(2009)]{minecraft}
{Mojang Studios}.
\newblock Minecraft, 2009.

\bibitem[Mokady et~al.(2023)Mokady, Hertz, Aberman, Pritch, and Cohen-Or]{mokady2023null}
Ron Mokady, Amir Hertz, Kfir Aberman, Yael Pritch, and Daniel Cohen-Or.
\newblock Null-text inversion for editing real images using guided diffusion models.
\newblock In \emph{CVPR}, 2023.

\bibitem[Poole et~al.(2023)Poole, Jain, Barron, and Mildenhall]{poole2022dreamfusion}
Ben Poole, Ajay Jain, Jonathan~T Barron, and Ben Mildenhall.
\newblock {DreamFusion}: Text-to-{3D} using {2D} diffusion.
\newblock In \emph{ICLR}, 2023.

\bibitem[Radford et~al.(2021)Radford, Kim, Hallacy, Ramesh, Goh, Agarwal, Sastry, Askell, Mishkin, Clark, et~al.]{radford2021learning}
Alec Radford, Jong~Wook Kim, Chris Hallacy, Aditya Ramesh, Gabriel Goh, Sandhini Agarwal, Girish Sastry, Amanda Askell, Pamela Mishkin, Jack Clark, et~al.
\newblock Learning transferable visual models from natural language supervision.
\newblock In \emph{ICML}, 2021.

\bibitem[Ren et~al.(2024)Ren, Liu, Zeng, Lin, Li, Cao, Chen, Huang, Chen, Yan, et~al.]{ren2024grounded}
Tianhe Ren, Shilong Liu, Ailing Zeng, Jing Lin, Kunchang Li, He Cao, Jiayu Chen, Xinyu Huang, Yukang Chen, Feng Yan, et~al.
\newblock Grounded {SAM}: Assembling open-world models for diverse visual tasks.
\newblock \emph{arXiv preprint arXiv:2401.14159}, 2024.

\bibitem[Rombach et~al.(2022)Rombach, Blattmann, Lorenz, Esser, and Ommer]{rombach2022high}
Robin Rombach, Andreas Blattmann, Dominik Lorenz, Patrick Esser, and Bj{\"o}rn Ommer.
\newblock High-resolution image synthesis with latent diffusion models.
\newblock In \emph{CVPR}, 2022.

\bibitem[Ruiz et~al.(2023)Ruiz, Li, Jampani, Pritch, Rubinstein, and Aberman]{ruiz2023dreambooth}
Nataniel Ruiz, Yuanzhen Li, Varun Jampani, Yael Pritch, Michael Rubinstein, and Kfir Aberman.
\newblock {DreamBooth}: Fine tuning text-to-image diffusion models for subject-driven generation.
\newblock In \emph{CVPR}, 2023.

\bibitem[Sharma et~al.(2018)Sharma, Goyal, Liu, Kalogerakis, and Maji]{sharma2018csgnet}
Gopal Sharma, Rishabh Goyal, Difan Liu, Evangelos Kalogerakis, and Subhransu Maji.
\newblock {CSGNet}: Neural shape parser for constructive solid geometry.
\newblock In \emph{CVPR}, 2018.

\bibitem[Shi et~al.(2024)Shi, Wang, Ye, Long, Li, and Yang]{shi2023mvdream}
Yichun Shi, Peng Wang, Jianglong Ye, Mai Long, Kejie Li, and Xiao Yang.
\newblock {MVDream}: Multi-view diffusion for {3D} generation.
\newblock In \emph{ICLR}, 2024.

\bibitem[Song et~al.(2021)Song, Meng, and Ermon]{song2020denoising}
Jiaming Song, Chenlin Meng, and Stefano Ermon.
\newblock Denoising diffusion implicit models.
\newblock In \emph{ICLR}, 2021.

\bibitem[Sun et~al.(2023)Sun, Han, Deng, Wang, Qin, and Gould]{sun20233d}
Chunyi Sun, Junlin Han, Weijian Deng, Xinlong Wang, Zishan Qin, and Stephen Gould.
\newblock {3D-GPT}: Procedural {3D} modeling with large language models.
\newblock \emph{arXiv preprint arXiv:2310.12945}, 2023.

\bibitem[Tam et~al.(2024)Tam, Pun, Wang, Chang, and Savva]{tam2024scenemotifcoder}
Hou In~Ivan Tam, Hou In~Derek Pun, Austin~T Wang, Angel~X Chang, and Manolis Savva.
\newblock {SceneMotifCoder}: Example-driven visual program learning for generating {3D} object arrangements.
\newblock \emph{arXiv preprint arXiv:2408.02211}, 2024.

\bibitem[Teed and Deng(2020)]{teed2020raft}
Zachary Teed and Jia Deng.
\newblock {RAFT}: Recurrent all-pairs field transforms for optical flow.
\newblock In \emph{ECCV}, 2020.

\bibitem[Tian et~al.(2019)Tian, Luo, Sun, Ellis, Freeman, Tenenbaum, and Wu]{tian2018learning}
Yonglong Tian, Andrew Luo, Xingyuan Sun, Kevin Ellis, William~T. Freeman, Joshua~B. Tenenbaum, and Jiajun Wu.
\newblock Learning to infer and execute {3D} shape programs.
\newblock In \emph{ICLR}, 2019.

\bibitem[Yamada et~al.(2024)Yamada, Chandu, Lin, Hessel, Yildirim, and Choi]{yamada2024l3go}
Yutaro Yamada, Khyathi Chandu, Yuchen Lin, Jack Hessel, Ilker Yildirim, and Yejin Choi.
\newblock {L3GO}: Language agents with chain-of-{3D}-thoughts for generating unconventional objects.
\newblock In \emph{ICLR Workshop}, 2024.

\bibitem[Yuille and Kersten(2006)]{yuille2006vision}
Alan Yuille and Daniel Kersten.
\newblock Vision as {Bayesian} inference: analysis by synthesis?
\newblock \emph{Trends in Cognitive Sciences}, 10\penalty0 (7):\penalty0 301--308, 2006.

\bibitem[Zhang et~al.(2023{\natexlab{a}})Zhang, Cai, Fu, Yuan, and Lu]{zhang2023creative}
Chi Zhang, Penglin Cai, Yuhui Fu, Haoqi Yuan, and Zongqing Lu.
\newblock Creative agents: Empowering agents with imagination for creative tasks.
\newblock \emph{arXiv preprint arXiv:2312.02519}, 2023{\natexlab{a}}.

\bibitem[Zhang et~al.(2023{\natexlab{b}})Zhang, Rao, and Agrawala]{zhang2023controlnet}
Lvmin Zhang, Anyi Rao, and Maneesh Agrawala.
\newblock Adding conditional control to text-to-image diffusion models.
\newblock In \emph{ICCV}, 2023{\natexlab{b}}.

\bibitem[Zhang et~al.(2018)Zhang, Isola, Efros, Shechtman, and Wang]{zhang2018unreasonable}
Richard Zhang, Phillip Isola, Alexei~A Efros, Eli Shechtman, and Oliver Wang.
\newblock The unreasonable effectiveness of deep features as a perceptual metric.
\newblock In \emph{CVPR}, 2018.

\bibitem[Zhou et~al.(2024{\natexlab{a}})Zhou, Li, Ma, Zhang, and Yang]{zhou2024migc}
Dewei Zhou, You Li, Fan Ma, Xiaoting Zhang, and Yi Yang.
\newblock {MIGC}: Multi-instance generation controller for text-to-image synthesis.
\newblock In \emph{CVPR}, 2024{\natexlab{a}}.

\bibitem[Zhou et~al.(2024{\natexlab{b}})Zhou, Hou, Luo, Wang, Zhang, and Peng]{zhou2024scenex}
Mengqi Zhou, Jun Hou, Chuanchen Luo, Yuxi Wang, Zhaoxiang Zhang, and Junran Peng.
\newblock {SceneX}: Procedural controllable large-scale scene generation via large-language models.
\newblock \emph{arXiv preprint arXiv:2403.15698}, 2024{\natexlab{b}}.

\bibitem[Zhou et~al.(2024{\natexlab{c}})Zhou, Ran, Xiong, He, Lin, Wang, Sun, and Yang]{zhou2024gala3d}
Xiaoyu Zhou, Xingjian Ran, Yajiao Xiong, Jinlin He, Zhiwei Lin, Yongtao Wang, Deqing Sun, and Ming-Hsuan Yang.
\newblock {GALA3D}: Towards text-to-{3D} complex scene generation via layout-guided generative {Gaussian} splatting.
\newblock In \emph{ICML}, 2024{\natexlab{c}}.

\bibitem[Zhu et~al.(2016)Zhu, Kr{\"a}henb{\"u}hl, Shechtman, and Efros]{zhu2016generative}
Jun-Yan Zhu, Philipp Kr{\"a}henb{\"u}hl, Eli Shechtman, and Alexei~A Efros.
\newblock Generative visual manipulation on the natural image manifold.
\newblock In \emph{ECCV}, 2016.

\end{thebibliography}
}

\newpage
\clearpage

\appendix
\section{Overview}
This file contains representation details (\cref{app:representation}), experiment details (\cref{app:exp}), extended experiment results (\cref{app:results}), discussions of limitations (\cref{app:failure}), and full language model prompts and outputs (\cref{app:lm}). 

\section{Representation Details}
\label{app:representation}

\subsection{Domain-Specific Language}
\label{app:dsl4d}

The complete DSL is listed in \cref{tab:dsl-v2-and-full}. We explain the four macros introduced in \cref{ssec:dsl}, also listed in \cref{tab:dsl-v2-and-full}, as follows. 
\begin{itemize}
    \item Macro \texttt{call} retrieves \texttt{<entity-func>} bound to the input word, applies the function on the input embeddings, and outputs a data object of type \texttt{Entity}. Specifically, \texttt{(car embedding-list)} is the embedding of the output entity corresponding to $z$ from \cref{eqn:Psi-f}, and \texttt{(cdr embedding-list)} is the embeddings of its descendent entities corresponding to $\gamma$. 
    \item Macro \texttt{union} composes transformed entities by aggregating inputs into a list.
    \item Macro \texttt{union-loop} applies \texttt{union} in a for loop.
    \item Macro \texttt{transform} pairs an entity with its pose. 
\end{itemize}

\begin{table*}[t]
\centering
\scriptsize
\setlength{\tabcolsep}{1pt}
\begin{tabular}{ll}
\toprule
 \multicolumn{2}{l}{Data Types} \\ 
\midrule
\texttt{\WordType} & 
\quad\texttt{//} \Word specifying semantics\\
\texttt{\EmbdType} & 
\quad\texttt{//} \Embd specifying an entity's attributes\\
\texttt{Matrix} & \texttt{::= Array[Array[Float]]}\quad\texttt{//} Transformation in $\mathrm{GA}(3, \mathbb{R})$ \\
\texttt{Entity} & \texttt{::= Tuple[Tuple[\WordType, \EmbdType\unskip], List[Tuple[Entity, Matrix]]]}
\\
\texttt{Vector} & \texttt{::= Array[Float]}\quad\texttt{//} Vector in $\mathbb{R}^3$ \\\midrule
 \multicolumn{2}{l}{Grammar} \\ 
\midrule
\texttt{<START>} &\texttt{::= \texttt{<bind-expr>*}}\\
\texttt{<bind-expr>} & \texttt{::= (bind <\word\unskip> <entity-func>)}\\
\texttt{<entity-func>} & \texttt{::= (lambda (\embd::\Embd embedding-list::List[\Embd\unskip])}\\
& \texttt{\quad\quad\quad\quad <sub-entities>)}\\
\texttt{<sub-entities>} & \texttt{::= (union <entity-transform>*)} \\
& \quad \texttt{| (union-loop <loop-count> (lambda (i::Integer) <entity-transform>))} \\
\texttt{<entity-transform>} 
& \texttt{::= (transform <entity> <matrix>)} \\
\texttt{<entity>} & \texttt{::= (call <\word\unskip> <\textcolor{myEmbdColor}{embedding}>*)} \\
\texttt{<\word\unskip>} & \texttt{:: \WordType} \\
\texttt{<entity-func>} & \texttt{:: \Embd~-> List[\Embd\unskip] -> Entity} \\
\texttt{<loop-count>} &\texttt{:: Integer} \\
\texttt{<matrix>} &\texttt{:: Matrix} \\
\texttt{<\embd\unskip>} & \texttt{:: \EmbdType} \\
\texttt{<4D-entity-func>} & \texttt{::= (lambda () <create-entity-list>)} \quad\texttt{//} Define a function that outputs a 4D scene\\
\texttt{<create-entity-list>} & \texttt{::= (list <entity>*)} \quad\texttt{//} Represent a 4D scene as a temporal list of entities \\
\midrule

 \multicolumn{2}{l}{Macros} \\ \midrule
\texttt{call} & \texttt{::= (lambda (\word~.~embedding-list)}\quad\texttt{//} Return an entity from the semantic class of \word\\
&\quad\quad\quad\quad\texttt{(cons (cons \word (car embedding-list))}\\
&\quad\quad\quad\quad\quad\quad\texttt{((retrieve \word) (car embedding-list) (cdr embedding-list))))}\\
\texttt{union} & \texttt{::= list}\quad\texttt{//} Compose transformed entities\\
\texttt{union-loop} & \texttt{::= (lambda (loop-count loop-func)} \quad\texttt{//} Compose transformed entities using a for loop \\ 
& \quad\quad\quad\quad\texttt{(union (map loop-func (iota loop-count))))} \\
\texttt{transform} & \texttt{::= cons}\quad\texttt{//} Transform entity pose\\
\texttt{call} & \texttt{:: \Word~-> List[\Embd\unskip] -> Entity} \\
\texttt{union} & \texttt{:: (Tuple[Entity, Matrix])* -> List[Tuple[Entity, Matrix]]} \\
\texttt{union-loop} & \texttt{:: Int -> (Int -> Tuple[Entity, Matrix])}\\
&\quad\quad\quad\quad\texttt{-> List[Tuple[Entity, Matrix]]]} \\
\texttt{transform} & \texttt{:: Entity -> Matrix -> Tuple[Entity, Matrix]} \\
\texttt{translate} & \texttt{:: Vector -> Matrix} \quad\texttt{//} Compute translation matrix \\
\texttt{rotate} & \texttt{:: Float -> Vector -> Vector -> Matrix} \quad\texttt{//} Compute rotation matrix\\
\texttt{scale} & \texttt{:: Vector -> Vector -> Matrix}\quad\texttt{//} Compute scaling matrix \\
\texttt{reflect} & \texttt{:: Vector -> Vector -> Matrix}\quad\texttt{//} Compute reflection matrix \\
\texttt{@} & \texttt{:: Matrix -> Matrix -> Matrix} \quad\texttt{//} Matrix multiplication\\
\texttt{compute-shape-center} & \texttt{::  Entity -> Vector}\quad\texttt{//} Compute center of an entity's bounding box\\
\texttt{compute-shape-min} & \texttt{::  Entity -> Vector}\quad\texttt{//} Compute minimum corner of an entity's bounding box\\
\texttt{compute-shape-max} & \texttt{::  Entity -> Vector} \quad\texttt{//} Compute maximum corner of an entity's bounding box \\
\texttt{compute-shape-sizes} & \texttt{::  Entity -> Vector}\quad\texttt{//} Compute sizes of an entity's bounding box\\
\midrule
\multicolumn{2}{l}{Special Forms} \\ 
\midrule
\multicolumn{2}{l}{\texttt{(bind <\word\unskip> <entity-func>)}\quad\texttt{//} Defines and binds an entity function} \\
\multicolumn{2}{l}{\texttt{(retrieve <\word\unskip>)} \quad\texttt{//} Retrieves an entity function bound to \word, or \texttt{(lambda (\_) (list))} if such function does not exist} \\
\bottomrule
\end{tabular}
\vspace{-0.7em}
\caption{
\textbf{The Domain-Specific Language.}
The table contains the DSL specification used to define our representation.
Built-in data types (\eg, \texttt{Float}), functions (\eg, \texttt{car} and \texttt{cdr}), special forms (\texttt{lambda}), and conditionals (\texttt{if}) are omitted; \texttt{<START>} denotes program starts, \texttt{::=} denotes rewriting rules; \texttt{::} denotes type annotation; \texttt{\_::} denotes type annotation for an anonymous formal parameter, \texttt{*} denotes one or more expressions of the same type. 
}
\vspace{-1.5em}
\label{tab:dsl-v2-and-full}
\end{table*}

\subsection{Details of Graphics Renderers}
\label{app:renderers}

This section expands the instantiation of three graphics renderers from \cref{ssec:render} in detail. For each renderer, we will discuss its parameter space $\Theta$ and $\mathcal{T}$, rendering operation $\mathcal{R}$, and the reparameterization function $g_\text{reparam}$. 

\subsubsection{SDS-Based Renderer}
\paragraph{Parameter Space with 3D Gaussians.}
For this renderer, $\Theta$ is the space of 3D Gaussian parameters and $\mathcal{T}$ is the space of 3D affine transformation matrices. 
In particular, each primitive is parameterized as a set of $K$ 3D Gaussians under a 3D affine transformation $t$, written as $(\theta, t) = (K, \{\phi_i\}_{i=1}^K, t) \in \Theta\times \mathcal{T}$, where $\phi_{i}$ is the set of parameters for a single 3D Gaussian, and $t$ is a 3D transformation matrix.
Each Gaussian parameter $\phi$ is defined as $\phi := (\mu, \alpha, s, q, c)$, denoting the 3D center position, opacity, scale, rotation in quaternion, and color of the Gaussian, respectively. 
An entity consisting of $N$ primitives is parameterized as $\{(\theta_j, t_j)\}_{j=1}^N = \{(K_j, \{\phi_{i}^j\}_{i=1}^{K_j}, t_j)\}_{j=1}^N$.

\paragraph{Differentiable Rendering.}
The rendering operation $\mathcal{R}$ for the 3D Gaussian renderer is as follows. 

Following \cite{kerbl3Dgaussians}, a single Gaussian is defined by 
\begin{equation*}
    G(x) = e^{-\frac{1}{2}(x-\mu)^T\Sigma^{-1}(x-\mu)},
\end{equation*}
where $x\in\mathbb{R}^3$ is a point in world coordinate, $\Sigma:=(RS)(RS)^T$ the 3D covariance matrix, $R$ the rotation matrix computed from $q$, and $S$ the scaling matrix computed from $s$. 

A Gaussian under transformation $t\in\mathcal{T}$ with $t(x) = {R_t}{S_t}x + p_t$, where $R_t, S_t, p_t$ are the rotation, scaling, and translation components, respectively, is then computed with $G_t$ satisfying the follows:
\begin{equation*}
    G_t(t(x)) = G(x).
\end{equation*}
We assume that diagonal entries of the scaling matrix $S_t$ are all positive, and therefore $t$ is invertible.
Solving for the above equation gives
\begin{equation*}
    G_t(x) = e^{-\frac{1}{2}(x - \mu_t)^T\Sigma_t^{-1}(x - \mu_t)}, 
\end{equation*}
where $\mu_t = t(\mu)$ and $\Sigma_t = ((R_tR)(S_tS))((R_tR)(S_tS))^T$. 
Let $\tilde{t}(\phi)$ be the Gaussian after applying transformation $t$ on $\phi$. Then $\tilde{t}(\phi)$ has center $\mu_t$, rotation $R_t R$, scale $S_t S$, and has $\alpha$ and $c$ remaining unchanged as derived above.

The rendering operation $\mathcal{R}$ to convert an entity consisting of $N$ primitives, $\{(\theta_j, t_j)\}_{j=1}^N = \{(K_j, \{\phi_{i}^j\}_{i=1}^{K_j}, t_j)\}_{j=1}^N$,
to the image space simply amounts to rendering all post-transformation 3D Gaussians in the scene, $\{\tilde{t}_j(\theta_j)\}_{j} := \{\tilde{t}_j(\phi_i)\}_{i,j}$, following the projection and blending process from \cite{kerbl3Dgaussians}.

\paragraph{Primitive Reparameterization via SDS Guidance.}
Recall that $g_\text{reparam}$ aims to obtain 3D Gaussian primitive parameters for per-primitive conditional embeddings $\{z_j\}_{j=1}^N$ and global condition $z_\text{global}$, where $z_j = g_\text{CLIP}(c_j)$ is explained in \cref{ssec:render}, and $z_\text{global} = g_\text{CLIP}(c_\text{global})$ is computed from a global scene description in texts, $c_\text{global}$. 
We now expand \cref{ssec:render} to describe the optimization process of $g_\text{reparam}$ in detail.

We write the SDS objective originally proposed in \cite{poole2022dreamfusion} as follows:
\begin{equation*}
\begin{aligned}
    &\nabla_\psi \mathcal{L}_\text{SDS}(x=\mathcal{R}(\psi); z, \hat{\epsilon}) \\
    = & \mathbb{E}_{\eta\sim\mathcal{U}(0, 1), \epsilon\sim\mathcal{N}(0, I)}\left[w(\eta)(\hat{\epsilon}(\alpha_\eta x + \alpha_\eta \epsilon, z, \eta) - \epsilon) \frac{\partial x}{\partial \psi}\right],
    \end{aligned}
\end{equation*}
where $\hat{\epsilon}$ is a pre-trained image denoising network, $\eta$ is diffusion timestep, and $w(\cdot), \alpha_\eta$ come from diffusion schedule. 

For entity $\{(\theta_j, t_j)\}_{j=1}^N$, let
\begin{equation*}
\begin{aligned}
    &\mathcal{L}(\{z_j\}_j, z_\text{global}, \{t_{\text{init},j}\}_j) \\
    := & \mathcal{L}_\text{SDS}(\mathcal{R}(\{\tilde{t}_j(\theta_j)\}_j); z_\text{global}, \hat{\epsilon}_\text{ControlNet}) \\
    & + \sum_j \mathcal{L}_\text{SDS}(\mathcal{R}(\theta_j);z_j, \hat{\epsilon}_\text{MVDream}) 
    \\
    &+ \sum_j\mathcal{L}_\text{reg}(\theta_j, \text{StopGrad}(t_j))
    + \sum_j \mathcal{L}_\text{layout}(\theta_j, t_{\text{init},j}),
\end{aligned}
\end{equation*}
where $\mathcal{L}_\text{reg}, \mathcal{L}_\text{layout}$ are regularization terms following the definition from \cite{zhou2024gala3d} and $\text{StopGrad}$ stops gradients from backpropagation. 
Here, $\mathcal{L}_\text{reg}$ penalizes Gaussian ellipsoids that are too long, and $\mathcal{L}_\text{layout}$ penalizes Gaussians that lie outside the intial bounding box specified by $t_{\text{init}}$.

Finally, we have
\begin{equation*}
    g_\text{reparam} = \argmin_{\{(\theta_j, t_j)\}_{j=1}^N}\mathcal{L}.
\end{equation*}

During optimization, if primitives $j_1$ and $j_2$ have the same condition and initial normalized bounding box scale, \ie, ($z_{j_1}=z_{j_2}) \wedge (\frac{S_{t_{j_1}}} {\|S_{t_{j_1}}\|_2} = \frac{S_{t_{j_2}}} {\|S_{t_{j_2}}\|_2})$, they are enforced to have the same parameters $\theta$ (but still distinct $t_{j_1}$ and $t_{j_2}$), which greatly reduces the number of parameters in the solution space.

In practice, for certain scenes, LM outputs treat detailed object parts as primitives, \eg, the hat rim and hat top from the first example in \cref{fig:app-exp-editing}, and the backbone model for SDS guidance cannot effectively model such fine-grained parts. Therefore, we treat the hat as a primitive, whose pose is computed from the minimum bounding box containing both the hat rim and hat top, before carrying out the above optimization. 
This process effectively adapts the granularity of the computation graph, originally specified in LM inference outputs, to the graphics renderer being used, by assigning intermediate nodes from the original computation graph as the new leaf nodes.

\subsubsection{Mitsuba Renderer}
\paragraph{Parameter Space.} 
For this renderer, $\Theta$ is the parameter space for three types of graphics primitives supported by Mitsuba: \texttt{cube}, \texttt{sphere}, and \texttt{cylinder}, as specified in the function header for \texttt{primitive\_call} in \cref{app:lm-prompt-generation}. $\mathcal{T}$ is the 3D affine transformation space. 

\paragraph{Renderer.}
We use the path tracer with maximum depth 8 implemented in Mitsuba. In this work, we use Mitsuba as a generic graphics engine and do not take advantage of its differentiability. 

\paragraph{Reparameterization.}
Since we directly prompt LM to generate Mitsuba primitive parameters in its outputs as specified in \cref{app:lm-prompt-generation}, the function parameters from raw LM outputs are already in the parameter space $\Theta$ and are directly used for rendering, instead of being encoded into CLIP embeddings $z\in \mathcal{Z}_\text{CLIP}$.

\subsubsection{Minecraft Renderer}

\paragraph{Parameter Space.} 
For this renderer, $\Theta$ is the asset parameters for Mincraft blocks, and $\mathcal{T}$ is the space of 3D similarity transformation matrices, \ie, of scaling and translation transformations. Note that we prevent rotation transformations in Minecraft, since that could lead to shapes that are impossible to render correctly in Minecraft.

Specifically, $\Theta$ is specified in the docstring from \cref{app:lm-prompt-minecraft} and is expanded below.
We introduce two types of primitives that let us construct in-game elements. 

The first is \texttt{set\_cuboid}. This primitive facilitates the creation of a cuboid within the Minecraft environment. The function accepts three arguments: (1) A string denoting the Minecraft block type (\eg, \texttt{"minecraft:white\_concrete"}); (2) A tuple of three integers representing the scaling along the x, y, and z axes; (3) A boolean flag, \texttt{fill}, that specifies whether the cuboid should be solid or hollow.
The cuboid is anchored at the coordinate origin $(0, 0, 0)$, which corresponds to its front-left-bottom vertex. 

The second is \texttt{delete\_blocks}. This primitive allows for the deletion of a previously placed cuboid. It accepts a single parameter, which is a tuple of three integers denoting the scaling along the x, y, and z axes.
This operation removes the cuboid with its front-left-bottom vertex at the origin $(0, 0, 0)$, effectively clearing the designated space.

Note that we do not provide the Minecraft block type in the prompt, but instead let the model choose this parameter. Since there is a large amount of Minecraft data files on the web, the model performs decently well in choosing appropriate Minecraft blocks. We also augment this by building safety checks; for example, if the model chooses a Minecraft block that doesn't exist in our version of Minecraft, we will use semantic similarity to choose the most similar block from our library.

\paragraph{Renderer.}
We use WebGL\footnote{\url{https://get.webgl.org/}} and Deepslate\footnote{\url{https://misode.github.io/deepslate/}} for rendering Minecraft builds.

\paragraph{Reparameterization.}
Similar to Mitsuba, function parameters from LM-generated programs are directly used for rendering without CLIP encoding or reparameterization.

\subsubsection{Text-to-Image (T2I) Model Renderer}

\paragraph{Parameter Space.} We employ MIGC~\citep{zhou2024migc} as the backbone model for this renderer, which originally uses a CLIP text encoder~\citep{radford2021learning} and a pre-trained UNet from Stable Diffusion~\citep{rombach2022high} for layout-conditioned text-to-image generation. The parameter space $\Theta$ for this renderer is the CLIP text embedding space.

\paragraph{Renderer.} We first project the 3D bounding boxes of primitives from an execution output of our representation to a 2D layout under a specified camera viewpoint, and then run the forward pass of the T2I model conditioned on the 2D layout, where each 2D bounding box corresponds to an aforementioned CLIP embedding $\theta\in\Theta$ . 

\paragraph{Reparameterization.} 
Function parameters from LM-generated programs are directly encoded by the CLIP text encoder using the language templates described in \cref{sec:inference}.

\section{Experiment Details}
\label{app:exp}

\subsection{Computation Cost}
On one scene, LM inference takes $<$1 min, primitives-based rendering takes $<$1 min, SDS-based rendering takes $\sim$30 min/object. 
All experiments run on 1 A5000 GPU with 48GB memory. 

\subsection{Textual Inversion Optimization}
\label{app:textual-inversion}
To obtain image-conditioned embedding, we follow the procedure proposed in \cite{gal2022textual}. For the input image $I$ and text prompt $c_j$, we first use $c_j$ as guidance of GroundingSAM~\cite{ren2024grounded} to obtain the desired mask of the corresponding entity. The cropped region is pad to square and resized to desired resolution, resulting in image target $I_j$. The background of $I_j$ is set to random grayscale color as used in \cite{shi2023mvdream}.

We adopt the language template \texttt{"<class>, 3d model, in the style of <style>"} in all the textual inversion experiments. The template is first converted into token embeddings, then using CLIP text-encoder $g_\text{CLIP}$ to transform to embeddings $z_j$ for diffusion model $\hat{\epsilon}_\text{MVDream}$. In each textual-inversion iteration, we optimize the token embeddings $v_{j1}, v_{j2}$ for \texttt{<class>} and \texttt{<style>} while freezing others. We use the similar objective as in diffusion model training:

\begin{equation*}
\begin{aligned}
&v_{j1}^*, v_{j2}^* = \\
&\argmin_{v_{j1}, v_{j2}} \mathbb{E}_{\eta, \epsilon}\Vert \epsilon - \hat{\epsilon}_\text{MVDream}(\alpha_{\eta}I_j + \alpha_{\eta} \epsilon,\eta, z_j(v_{j1}, v_{j2})) \Vert_{2}^{2}.
\end{aligned}
\end{equation*}

For each entity, we optimize the corresponding embeddings for $100$ iterations with learning rate $1$e-$2$. Empirically we find this setting is enough to fit the image conditions. After textual inversion, the embedding $z_j$ is computed with optimized token embeddings, and used to guide the entity optimization as explained in \cref{app:renderers}.

\subsection{GraphDreamer Implementation}
\label{app:graphdreamer}
Since the original paper didn't release the script for automatic scene graph generation, we follow the descriptions in the paper and re-implement this stage to query LM to output scene graphs in json format to avoid manually converting LM outputs to model configurations. 
The full system prompt is shown below:
\styledmarkdowninput{programs/system_prompt_graphdreamer.md}
The full user prompt is shown below, where the given example input and output are taken from the teaser figure of the original paper~\citep{gao2024graphdreamer}. 
In below, \texttt{\{task\}} is a placeholder for input text prompts of scenes.
\styledmarkdowninput{programs/user_prompt_graphdreamer.md}

After generating scene graphs with aforementioned prompt, we follow the released implementation from GraphDreamer to optimize for the final 3D representation, where each object in the node list is represented as an individual object SDF and color field.
SDS loss is used to optimize each object and object pairs described in the edge list. 
When the raw scene graph output contains too many objects and exceeds the memory limitation required in the optimization, we rerun the graph generation step and add ``The maximum number of objects is three'' in the text prompt and rerun the optimization. 

\begin{figure*}[t]
  \centering
  \includegraphics[width=\linewidth]{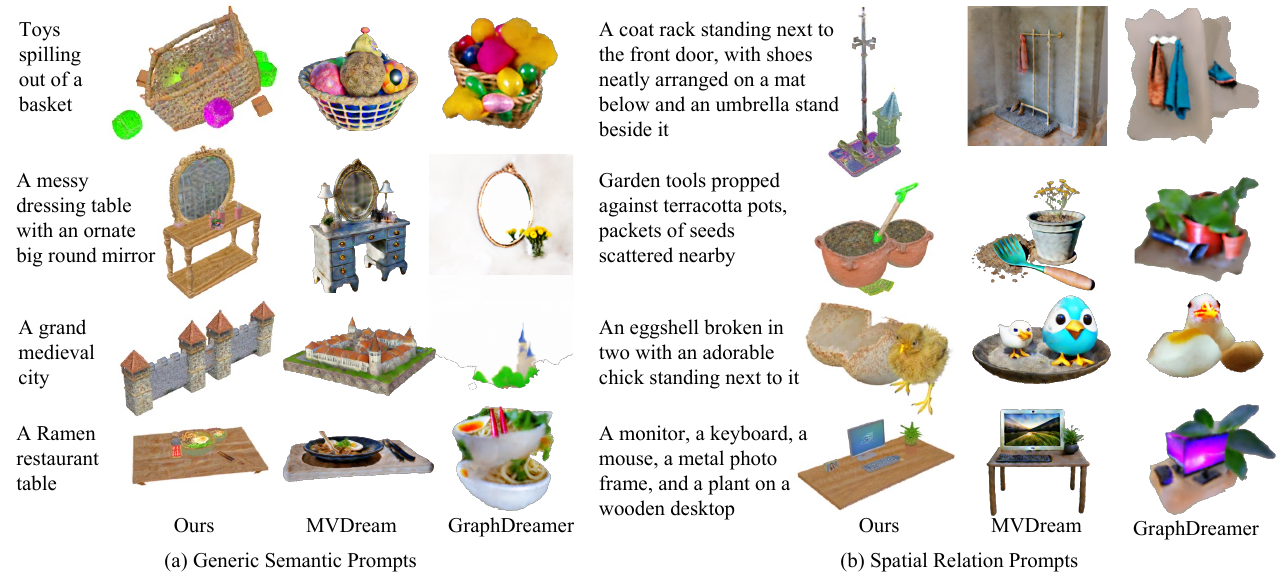}
  \vspace{-2em}
   \caption{\textbf{Text-Prompted 3D Generation Comparisons} extending \cref{fig:generation}.}
   \label{fig:app-generation}
\end{figure*}

\begin{table}[t]
\scriptsize
\setlength{\tabcolsep}{2pt}
    \centering
    \begin{tabular}{lcc}
        \toprule
        Methods & 
        CLIP~\citep{ilharco_gabriel_2021_5143773} ($\uparrow$)
        & Dynamic~\citep{huang2024vbench} ($\uparrow$) \\ \midrule
        4D-fy~\citep{bahmani20244dfy} & 0.352 & $0.2\%$ \\
        4D-fy~\citep{bahmani20244dfy} (w/ prompt variant) & \textbf{0.354} & $0.6\%$ \\
        Ours & 0.341 & $\textbf{5.9\%}$ \\ \bottomrule
    \end{tabular}
    \caption{\textbf{Evaluation for Text-Prompted 4D Generation} extending \cref{tab:lpips-flow}. 
    }
    \label{supp:4d}
\end{table}

\begin{figure*}[t]
    \centering
    \includegraphics[width=\linewidth]{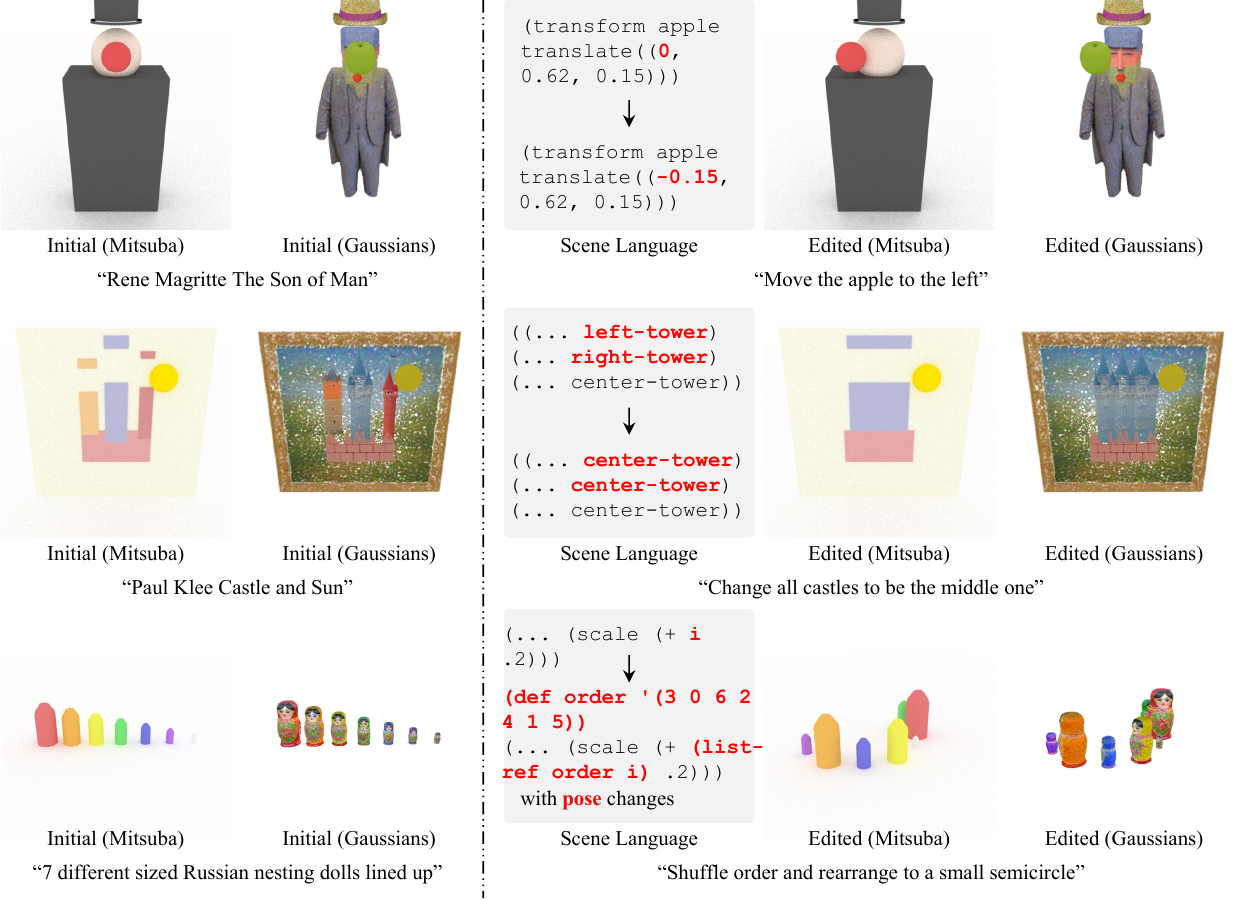}
    \vspace{-2em}
    \caption{\textbf{Scene Editing with Language Instructions} extending \cref{fig:editing}.}
    \label{fig:app-exp-editing}
\end{figure*}
\begin{figure}[t]
    \centering
    \includegraphics[width=0.6\linewidth]{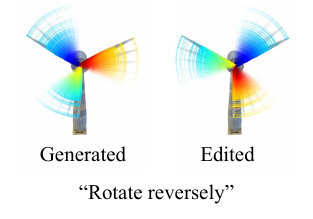}
    \vspace{-1em}
    \caption{\textbf{4D Editing.}}
    \label{fig:app-exp-editing-4d}
\end{figure}

\subsection{Evaluation Metrics}
\label{app:metrics}
\subsubsection{User Study}
\label{app:user-study}

The user study is conducted on Prolific\footnote{\url{https://www.prolific.com/}} with $103$ users. 
We provide the following instruction in the user study: 
``In the following questions, you'll be provided a list of text descriptions and corresponding three generation results. Please choose the one that has best text-alignment, taking into account both the visual quality and the adherence to the layout structure outlined in the text prompt.''
For each of the $9$ scenes being evaluated, we ask the user to choose one of the three video rendering results, generated by our method and two baselines in randomized order. All videos shown in the user study are displayed on the project page.

\subsubsection{CLIP-Based Evaluation}
\label{app:clip}
We use OpenCLIP model for measuring image-text similarity (\cref{tab:user-study,tab:lpips-flow}). 
We use the model variant with the highest ImageNet zero-shot classification accuracy, OpenCLIP-ViT-H-14-378-quickgelu. 

\begin{figure*}[t]
  \centering
  \includegraphics[width=\linewidth]{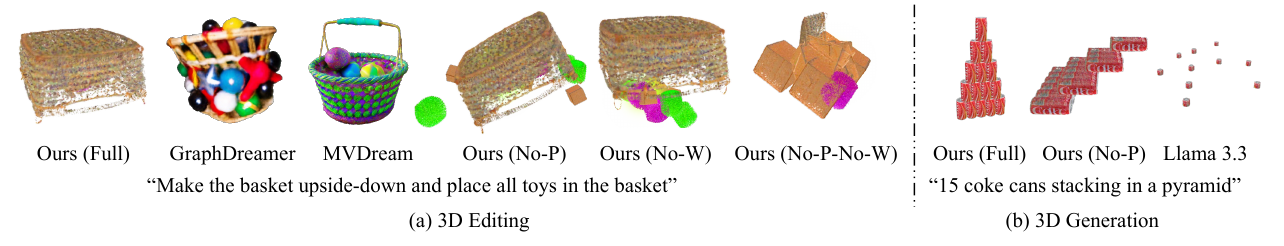}
  \vspace{-2em}
   \caption{\textbf{Ablations.}}
   \label{fig:app-ablations}
   \vspace{-1em}
\end{figure*}

\begin{figure}[t]
        \centering
 \includegraphics[width=\linewidth]{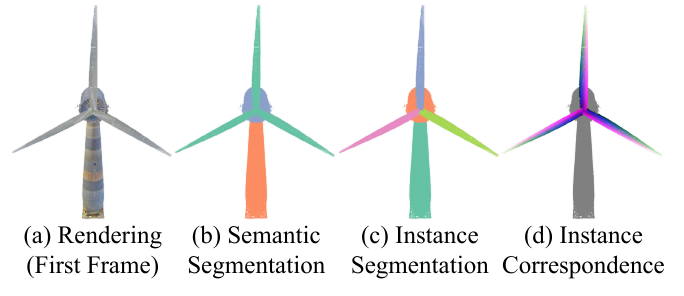}
        \vspace{-2.0em}
    \caption{\textbf{Visualizations of Discriminative Maps.} 
    }
    \label{fig:discriminative}
\end{figure}
\begin{figure}[t]
        \centering
 \includegraphics[width=\linewidth]{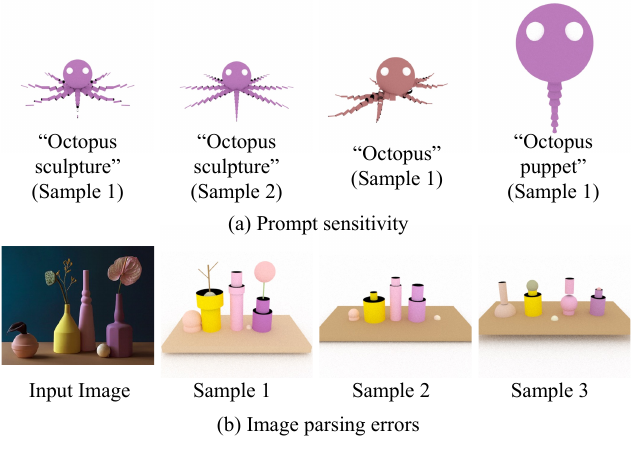}
    \caption{\textbf{Failure Cases.} }
    \label{fig:app-failure}
\end{figure}
\section{Extended Experiment Results}
\label{app:results}

\subsection{Text-Prompted 3D Scene Generation.}
\label{app:exp-generation}

Qualitative examples for numeric scenes and generic scenes are shown on the website and in \cref{fig:app-generation}, respectively. 

\subsection{Text-Prompted 4D Scene Generation}
\label{app:exp-4d}

\paragraph{Representation.}
The 4D entity function mentioned in \cref{ssec:text-cond-generation} is defined as \texttt{<4D-entity-func>} in the DSL defined in \cref{tab:dsl-v2-and-full}. 

\paragraph{Analysis.}
Allowing for a flexible set of primitive entities is crucial to make our representation suitable for generating 4D scenes of different scales, including objects with moving parts (\eg, the wind turbine from \cref{fig:generation4d}) and scenes with moving objects (\eg, the carousel). This is in contrast to prior works using primitives chosen from a fixed set~\citep{tian2018learning,sharma2018csgnet} or fixing the granularity to be object-centric~\citep{johnson2015image}. 

Moreover, the hierarchical scene structure encapsulated by our program-based representation makes it possible to represent 4D scenes compactly, serving as a regularization for generation output. Multiple entities, such as outputs from the function \texttt{horse} from the carousel scene in \cref{fig:generation4d}, can be grouped into one function \texttt{horses} and thereby share the same temporal transformation.
Writing composible functions for entity grouping effectively reduces the dimension of the temporal motion space and improves motion fidelity. 

\paragraph{Extended Quantitative Results.}
On top of the evaluation in \cref{tab:lpips-flow} with the same set of prompts for 4D-fy and ours, we evaluate 4D-fy with prompt variants that more explicitly indicate scene dynamics. Specifically, we use the following three prompts:
"a solar system model with moving planets",  "a toy wind turbine with rotating blades", and "carousel with a small canopy rotating". 
Results are reported as 4D-fy (w/ prompt variants) in \cref{supp:4d}. We observe a small increment in CLIP similarity and dynamic degrees in 4D-fy results with the prompt variants, but still with a relatively small motion compared to ours.

\subsection{Text-Prompted 3D and 4D Scene Editing}
\label{app:exp-editing}
\cref{fig:app-exp-editing} shows further qualitative experiment results under the same setup as \cref{fig:editing}. The same method applies to 4D editing, with results included in \cref{fig:app-exp-editing-4d}.

\subsection{Roles of Representation Components}  
Below we discuss the roles of representation components, $P$ (programs), $W$ (words), and $Z$ (embeddings).

We ablate the roles of $P$ and $W$ on the 3D editing task as follows: (i) \textit{No-$P$}, replacing the scene language with a leaf entity list (\ie, a list of word-pose tuples) before querying language models (LMs) to edit, (ii) \textit{No-$W$}, replacing function names with random strings, and (iii) \textit{No-$P$-No-$W$}, removing words from (i). 

We run these variants, together with baseline methods, to edit the top-left scene in \cref{fig:app-generation}. Input prompts and results are shown in \cref{fig:app-ablations}(a). 
Removing $P$ and/or $W$ degrades the accuracy of the results, suggesting that both components contribute to an intuitive and accurate editing interface. 
On the other hand, the two baseline methods do not encode precise spatial information of scene components and require regenerating the scene (we use prompt ``A basket upside-down with toys in the basket''), failing to preserve the original generated content. 

Embeddings $Z$ are necessary for image-prompted editing (\cref{fig:style_transfer}). 
For generation tasks, $W$ and $Z$ are both required by neural renderers, and we ablate $P$ by querying LMs to directly generate entity lists. As shown in \cref{fig:app-ablations}(b), removing programs harms LM accuracy; alternative backbone Llama has inferior performance.

\subsection{Visualization of Discriminative Information}
\label{app:discriminative}

Several pieces of discriminative information can be directly obtained with the proposed \reprFull: semantic maps in \cref{fig:discriminative}(b), as words represent per-entity semantics; instance segmentation in \cref{fig:discriminative}(c), as the representation is compositional with separable instances; correspondence of the repeated instances in \cref{fig:discriminative}(d), as programs specify repetitions existing in a scene; dense temporal correspondence for 4D scenes, as shown in \cref{fig:generation4d}. 

\section{Limitations}
\label{app:failure}
The current inference pipeline inherits limitations of the backbone pre-trained models. 

\paragraph{LM/VLM errors.} For text-conditioned tasks, minor variations in textual scene descriptions can lead to large quality differences in the output as shown in \cref{fig:app-failure} (a). 
For image-conditioned tasks, input images are parsed with the backbone visual language model. In the example in \cref{fig:app-failure} (b), with the same input image, parsing results have high variance across multiple inference runs.

\paragraph{T2I Model Bias.}
While the non-neural engines adopted are fully controllable and correct, neural-based engines may introduce additional errors, \eg, the rendered umbrella is not fully folded (\cref{fig:app-generation}) due to the bias in T2I models. 
Furthermore, the controls over texture and shapes are not disentangled for neural engines, resulting in mixed texture and shape changes in \cref{fig:style_transfer} as they are both affected by embeddings. 

While this work provides a viable inference method for the proposed representation that leverages the commonsense knowledge and code-writing capability of LMs and expressive renderings from T2I models, addressing the weaknesses inherited from these pre-trained models would further improve the robustness, complexity, and output quality for downstream tasks. We leave these as future directions to improve the inference of the \reprFull.

\section{Language Model Prompts and Responses}
\label{app:lm}

\subsection{Language Model Prompts}
\label{app:lm-prompt}

\subsubsection{Text- and Image-Conditioned Scene Generation}
\label{app:lm-prompt-generation}
In \cref{sec:inference}, we introduced an inference method for the representation by prompting LMs. 
The full system prompt is displayed below. The system prompt defines the data types and the function headers of macros from the DSL in \cref{tab:dsl-v2-and-full}, written in Python.
\styledmarkdowninput{programs/system_prompt.md}

The full user prompt for image or text-conditioned 3D generation is displayed below. It includes an example valid program, and the task specification indicated with a placeholder \texttt{\{task\}}. 
For text-conditioned generation (\cref{ssec:text-cond-generation}), it is replaced with the input textual scene description. 
For image-conditioned generation (\cref{ssec:image-cond-generation}), it is replaced with \texttt{"Reconstruct the input scene"}, and the input image is also fed into LM.
\styledmarkdowninput{programs/user_prompt.md}

\subsubsection{Scene Editing}

For text-prompted scene editing (\cref{ssec:text-cond-generation}), we prompt the LM in two rounds, first with a textual scene description with the same protocol from \cref{ssec:text-cond-generation}, and then with an editing instruction, \eg, \texttt{"move the apple to the left"}. In the second round, the system prompt remains the same as \cref{app:lm-prompt-generation}. The user prompt is as follows, where \texttt{\{program\}} is the LM output from first round, and \texttt{\{task\}} is the editing instruction. 

\styledmarkdowninput{programs/user_prompt_edit.md}

\subsubsection{4D Generation}
For 4D generation, we include one more macro definition in the system prompt as shown below, with the remaining system prompt the same as above:
\styledmarkdowninput{programs/system_prompt_4d.md}

The full user prompt for 4D generation is displayed below:

\styledmarkdowninput{programs/user_prompt_4d.md}

\subsubsection{Minecraft Rendering}
\label{app:lm-prompt-minecraft}

We adapt the prompts from \cref{app:lm-prompt-generation} to prompt LM to generate Minecraft-compatible outputs. Note that programs generated in this section can still be rendered using Mitsuba renderer, as shown in \cref{fig:aligned}, by converting each Minecraft block from LM outputs to a Mitsuba \texttt{cube} primitive with the corresponding location and with the color set as the average color of the Minecraft block.

Specifically, we remove \texttt{rotation\_matrix} and \texttt{reflection\_matrix} from the system prompt in \cref{app:lm-prompt-generation} and change the function header for \texttt{primitive\_call} to the follows: 
\styledmarkdowninput{programs/system_prompt_minecraft.md}

And we change the example program for user prompt accordingly to the follows:
\styledmarkdowninput{programs/user_prompt_minecraft.md}

\subsection{Language Model Outputs}
\label{app:lm-output}

\subsubsection{Text-Conditioned 3D Generation}
\label{app:lm-output-text-cond}
\paragraph{Example 1.}
This example corresponds to \cref{fig:teaser}. 

Input: \texttt{"Moai in Ahu Akivi, with slight variations"}.

Output:
\styledmarkdowninput{programs/outputs/moai.txt}

\paragraph{Example 2.}
This example corresponds to \cref{fig:representation}. 

Input: \texttt{"a chessboard at game start"}. 

Output:
\styledmarkdowninput{programs/outputs/chessboard.txt}

\subsubsection{Text-Conditioned 3D Editing}
In this task, for each example, we prompt LM in two rounds---the first round with a scene description as task specification, and the second round with editing instructions. LM outputs in the first round are omitted. Rendering outputs are shown in \cref{fig:editing} and \cref{fig:app-exp-editing}.

\paragraph{Example 1.}
Input scene description: \texttt{"Bramante Staircase, Vatican Museums"}. 
Input editing instruction: \texttt{"shrink staircase radius by 80\%"}. 

Output:
\styledmarkdowninput{programs/outputs/edited/staircase.txt}

\paragraph{Example 2.}
Input scene description: \texttt{"a 8-layer 3-color Jenga set at game start"}. 
Input editing instruction: \texttt{"remove 2 blocks from second top layer and the tower should not fall"}. 

Output:
\styledmarkdowninput{programs/outputs/edited/jenga.txt}

\paragraph{Example 3.}
Input scene description: \texttt{"a fractal tree-like structure"}. 
Input editing instruction: \texttt{"make branching structure to be trinary and 3D"}. 

Output:
\styledmarkdowninput{programs/outputs/edited/fractal.txt}

\paragraph{Example 4.}
Input scene description: \texttt{"Rene Magritte The Son of Man"}. 
Input editing instruction: \texttt{"move the apple to the left"}. 

Output:
\styledmarkdowninput{programs/outputs/edited/apple-man.txt}

\paragraph{Example 5.}
Input scene description: \texttt{"Paul Klee Castle and Sun"}. 
Input editing instruction: \texttt{"change all castles to be the middle one"}. 

Output:
\styledmarkdowninput{programs/outputs/edited/castle.txt}

\paragraph{Example 6.}
Input scene description: \texttt{"7 different sized Russian nesting dolls lined up"}. 
Input editing instruction: \texttt{"Shuffle order and rearrange to a small semicircle"}. 

Output:
\styledmarkdowninput{programs/outputs/edited/dolls.txt}

\subsubsection{Text-Conditioned 4D Generation}
\label{app:lm-output-4d}
\paragraph{Example 1.}
Input: \texttt{"carousel with a small canopy"}.

Output:
\styledmarkdowninput{programs/outputs/carousel.txt}
\paragraph{Example 2.}
Input: \texttt{"solar system model"}.

Output:
\styledmarkdowninput{programs/outputs/solar.txt}
\paragraph{Example 3.}
Input: \texttt{"a toy wind turbine"}.

Output:
\styledmarkdowninput{programs/outputs/wind.txt}

\end{document}